\documentclass[pdflatex,sn-mathphys-num]{sn-jnl}

\usepackage{graphicx}%
\usepackage{multirow}%
\usepackage{amsmath,amssymb,amsfonts}%
\usepackage{amsthm}%
\usepackage{mathrsfs}%
\usepackage[title]{appendix}%
\usepackage{xcolor}%
\usepackage{textcomp}%
\usepackage{manyfoot}%
\usepackage{booktabs}%
\usepackage{algorithm}%
\usepackage{algorithmicx}%
\usepackage{algpseudocode}%
\usepackage{listings}%
\usepackage{longtable}
\usepackage{rotating}

\theoremstyle{thmstyleone}%
\theoremstyle{thmstyletwo}%

\theoremstyle{thmstylethree}%

\begin{document}

\title{Transforming CCTV cameras into NO$_2$ sensors at city scale for adaptive policymaking}



\author[a,b,*]{Mohamed R. Ibrahim}
\author[a,c]{Terry Lyons}

\affil[a]{The Alan Turing Institute, London, UK}
\affil[b]{Department of Environment, University of Leeds, Leeds, UK}
\affil[c]{Mathematical Institute, Oxford University, Oxford, UK}
\affil[*]{Corresponding author, email: geomi@leeds.ac.uk}



\abstract{Air pollution in cities, especially NO\textsubscript{2}, is linked to numerous health problems, ranging from mortality to mental health challenges and attention deficits in children. While cities globally have initiated policies to curtail emissions, real-time monitoring remains challenging due to limited environmental sensors and their inconsistent distribution. This gap hinders the creation of adaptive urban policies that respond to the sequence of events and daily activities affecting pollution in cities. Here, we demonstrate how city CCTV cameras can act as a pseudo-NO\textsubscript{2} sensors. Using a predictive graph deep model, we utilised traffic flow from London's cameras in addition to environmental and spatial factors, generating NO\textsubscript{2} predictions from over 133 million frames. Our analysis of London's mobility patterns unveiled critical spatiotemporal connections, showing how specific traffic patterns affect NO\textsubscript{2} levels, sometimes with temporal lags of up to 6 hours. For instance, if trucks only drive at night, their effects on NO\textsubscript{2} levels are most likely to be seen in the morning when people commute. These findings cast doubt on the efficacy of some of the urban policies currently being implemented to reduce pollution. By leveraging existing camera infrastructure and our introduced methods, city planners and policymakers could cost-effectively monitor and mitigate the impact of NO\textsubscript{2} and other pollutants.}




\maketitle

\section{Introduction}\label{sec1}

Cities house more than half of the world's population \cite{source9}, which influence individuals’ behaviour \cite{source10} as well as their physical \cite{source11,source12} and mental health \cite{source13}. Every day, hundreds of millions of people spend several hours commuting on the spatial network of cities exposed to several risks, including air pollution. There is no dispute about the need for developing a fundamental understanding of how, collectively, individuals move from one location to another in their daily lives. This could be linked with pollution indicators to aid in emission reduction.

Nitrogen dioxide (NO\textsubscript{2}) is a major pollutant that can harm severely one's health \cite{source1,source2,source3,source4,source14,source15,source16}. NO\textsubscript{2} is formed by the combustion of fuels such as natural gas, diesel, petrol, and coal, and it can be found in the air as a result of traffic or a variety of land uses in cities, including industrial processes. NO\textsubscript{2} levels (measured in $\mu g/m^3$) vary in major cities worldwide \cite{source17}. Several studies have mapped NO\textsubscript{2} emissions from space \cite{source17,source18,source19,source20,source21,source22,source23,source24,source25}, whether during pandemics \cite{source17, source26} or after a policy is implemented  \cite{source18, source20, source24, source27}. While relying on satellite imagery is beneficial for many cases, including understanding the change in emission over a long period or across several large cities \cite{source19, source20, source24, source26, source28}, the spatial and temporal representations are often limited for understanding the dynamics of emission at a neighbourhood, district, or even many of the cities globally. Consequently, a substantial knowledge gap exists in linking micro-level events occurring frequently to their impact on emissions, thereby hindering the ability of policymakers to take localised actions. The objectives of this study are as follows: 1) to what extent the existence of specific traffic modes influences the surface NO\textsubscript{2} level, 2) what effect congestion and stationary modes have on the level of NO\textsubscript{2}, and 3) whether there is a significant temporal lag between what happens in traffic now and its impact on the future level of NO\textsubscript{2} at a given location. 

Analysing urban dynamics at the street level through visual data can uncover details that may be missed when when observing from space \cite{source29}. Recent progress in deep learning for predicting traffic flow \cite{10143389} aids in estimating pollutant levels in cities. Multi-modal sensor fusion has advanced by integrating data from various sensors to improve environmental predictions \cite{ma2023reducing}. These techniques could enable air quality estimation by combining CCTV visuals with other sensor data. Effective sensor deployment is crucial for urban-scale monitoring to ensure comprehensive coverage and reliable data collection \cite{song2022toward}. In this study, we introduce innovative techniques that leverage statistical analysis and graph neural networks to sense ambient ground-level NO\textsubscript{2} concentrations and their underlying factors using CCTV camera feeds on a citywide scale. This approach proves invaluable, especially in cities lacking an extensive network of environmental sensors. It provides an automated means of detecting the concentration of NO\textsubscript{2} levels and their causes related to the dynamics of traffic, empowering urban planners, and policymakers to actively monitor and respond to emerging issues in real-time, guided by the dynamic flow patterns within cities. Our methodology offers a non-physical (hardware-free) solution for monitoring ground-level NO\textsubscript{2} in urban areas where CCTV cameras are prevalent but NO\textsubscript{2} sensors are scarce, a situation encountered in numerous cities worldwide.

 \begin{figure*}
\centering
\includegraphics[width=1\linewidth]{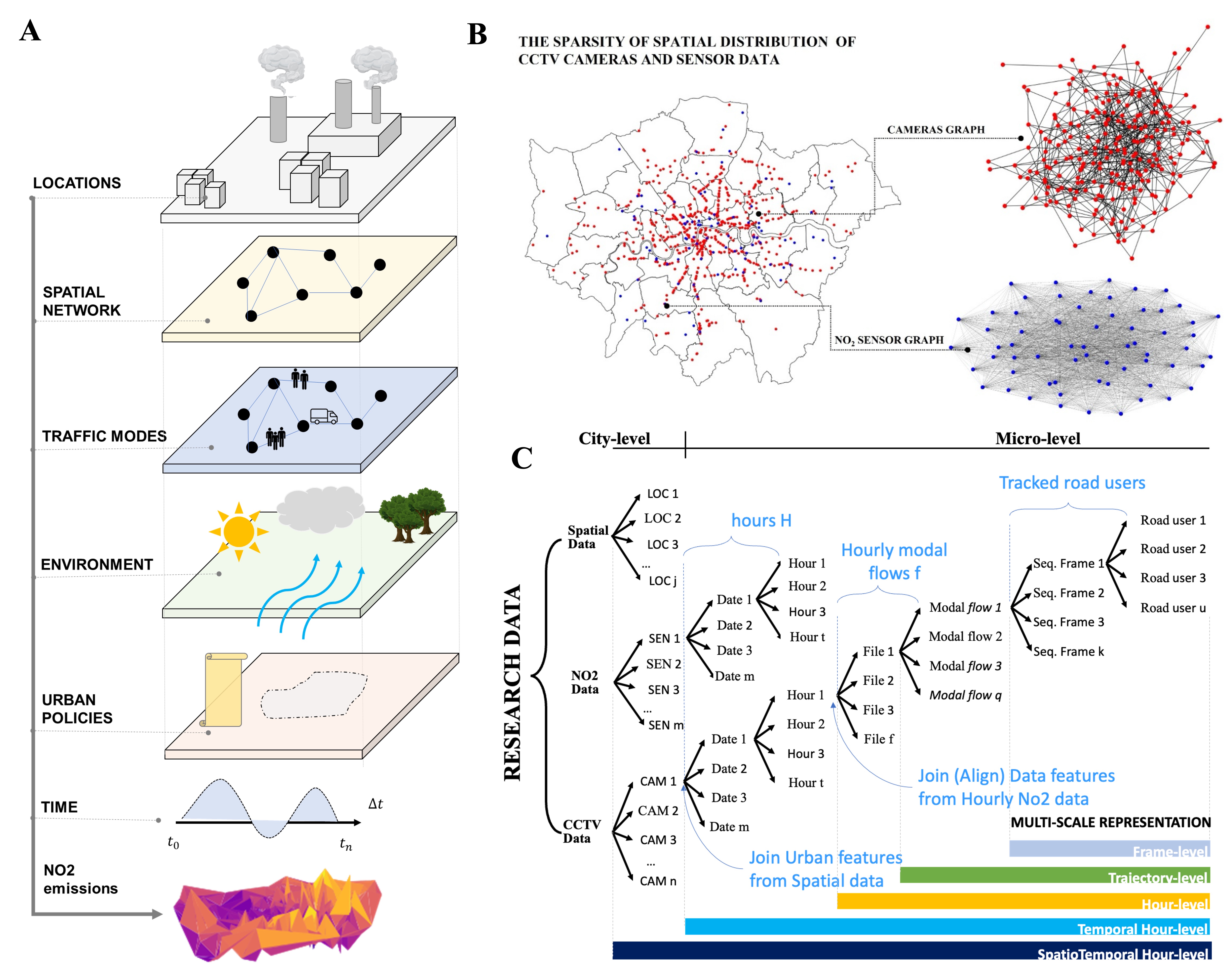}

\caption{Multi-level representation of all data sources. \textbf{(A)} The six layers of factors presented in this research. \textbf{(B)} The spatial representation as graph of knowledge of the camera (red nodes) and NO\textsubscript{2} (blue nodes) inputs. The locations of the cameras and NO\textsubscript{2} sensors do not need to align.\textbf{(C)} The multi-level representation of the studied data modalities shows several spatial and temporal resolutions in which different data modalities are aligned to conduct this research. }
\label{fig:fig1}
\end{figure*}

\begin{figure*}
\centering
\includegraphics[width=1\linewidth]{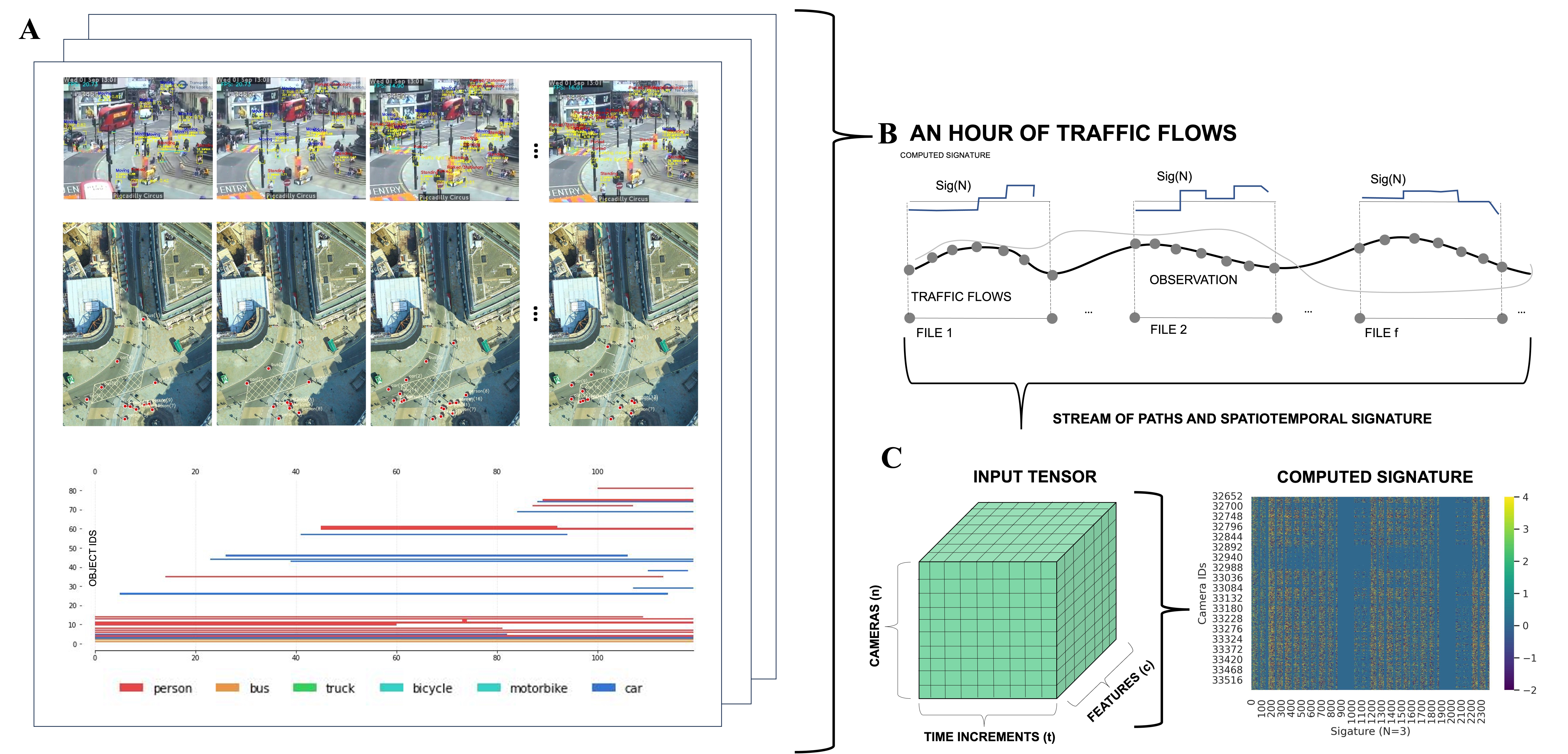}
\caption{Capturing the order of events from micro-level to city-scale \textbf{(A)} Shows 1) sequential frames of a given video file at Piccadilly circus in London as an input of one given CCTV camera, 2) vector representation of road users in an estimated bird’s eye view map with Google Maps to validate the geographical localisations of road users, and 3) temporal representation of road users within a given file based on the tracked system. \textbf{(B)} The relationship between hourly observed traffic flow data and the unseen temporal intervals among various file increments representing the stream of paths $X \in \mathbb{R}^{(nXtXc)}$, given that n is the number of cameras $(n=907)$, t is the number of file increments that make an hour of traffic modes $(t=11)$ and $c$ is the number of channels for traffic modal flows and their stationary status $(c=13)$ \textbf{(C)} The tensor representation of the generated paths with all its channel and their unique computed signatures ($Sig^N,N=3$)  that summarise the paths of varied traffic modal flows and their actions in a given scene. }
\label{fig:fig2}
\end{figure*}

\section{Results}\label{sec2}

\subsection{Multi-level Spatiotemporal representation of traffic modes}
To understand the influence of individual road users and their transportation modes on NO\textsubscript{2} ground-levels within the city, adopting a bottom-up approach that details individual trajectories is crucial. This strategy is invaluable for accurately assessing the real-time NO\textsubscript{2} concentrations at specific locations and times, as well as evaluating the exposure that individuals face during their commutes. Previous research across various domains has explored the use of human trajectories from GPS data for similar assessments \cite{gps1,gps2,gps3,gps4}. However, the limited availability of such data and substantial privacy concerns complicate the widespread replication of these methods. Therefore, it is imperative to discover alternative data sources that can accurately reflect traffic dynamics and roadway user behaviours while preserving anonymity. Successfully identifying such sources is key to advancing this study and enabling its future application across global urban landscapes to enhance our understanding of ground-level NO\textsubscript{2} distributions and their impacts on public health.

We used an open-access video data set provided by Transport For London (TfL), which includes unidentifiable human subjects and road users \cite{source30}. We recorded and analysed 133,132,866 sequential frames representing 112 unique hours in 907 London locations. We recorded many features of road users by utilising deep learning in our proposed framework. We refers to 'flows' as the movement patterns of road users captured by CCTV cameras across different locations and times within the city. These flows represent the dynamic interactions and traffic patterns, identified through the analysis of sequential frames in video data. By "flows," we mean the aggregated and continuous movement of vehicles and pedestrians detected and tracked through video footage. This term encompasses both the spatial and temporal dimensions of traffic, enabling us to infer NO\textsubscript{2} levels from the volume and behaviour of traffic over given periods.

Figure \ref{fig:fig1} illustrates the variables analysed and the structured hierarchy used to represent data for this study's various components. The data aims to depict diverse events and aspects of urban environments across different spatial and temporal scales (Fig. \ref{fig:fig1}-A, \ref{fig:fig1}-C). For example, the spatial distribution of data derived from camera streams does not necessarily match the spatial distribution of NO\textsubscript{2} sensors (Fig. \ref{fig:fig1}-B). Additionally, the temporal characteristics of data sourced from cameras, static spatial features, and NO\textsubscript{2} measurements differ (Fig. \ref{fig:fig1}-C).
At a micro-level of a given street, we extracted road users which were given a unique ID across the frame sequence of a given video file of a time increment of a given hour. Afterwards, a unique traffic modal flow (o) for a given hour is defined as   where $q$ is the different modal flows and $F$ is the number of different video files representing time increments of a given hour.  At a city scale, the data is combined for each unique hour (H)  of a given date (d) and hour (t). The overall Spatiotemporal representations of the CCTV data (X) is structured as $X \in \mathbb{R}^{H X N X F X C}$ and the generated NO\textsubscript{2} (Y) as $Y \in \mathbb{R}^{H X M}$, where H is the number of unique hours, N is the number of cameras’ locations, C is the number of features, including modal flows and locational urban features, and M is the number of NO\textsubscript{2} sensors’ locations where $M \neq N$. The spatiotemporal representations of cameras’ data and NO\textsubscript{2} sensors differ in position and temporal resolution, and they are aligned based on the sparse availability at hourly rates of NO\textsubscript{2} sensor data. The static urban features of a specific site are combined with the aligned locations of both sensor data. Time resolution remains as a variable depending on a given scale; moving from 0.04 sec at a frame level to 4 min in a trajectory level and finally to one hour at an aggregate higher level.  The construction of a non-linear tree data structure allows for the insertion, search, and relocation of new branches over time. It also supports this research by responding to stated questions that may require different spatial and temporal resolutions.

\subsection{Traffic composition at micro-scale}

To address how we can use high-frequency data (0.04 sec) of the number of road users and their behaviour (moving, stationary, etc.) to provide meaningful statements for NO\textsubscript{2} at an hourly city level, we must first collect and understand the collective patterns of road users at a micro-level that derive the overall traffic in London. We demonstrate, in Fig. \ref{fig:fig2},  how to transform the sequential frames of a given video to spatial and temporal representations of road users, and georeferencing their representation in a bird’s eye view map blended with Google map. We determined the modal flows based on the monitored unique ids of road users through the length of a given file to avoid re-counting the same users (Fig. \ref{fig:fig2}-B).  Lastly, to provide a unique summary of the observed sequence of the events of multidimensional streams of road users based on their types and behaviour at a given camera, we computed a signature, based on rough path theory \cite{source31, source32,source33}, \( \text{Sig}^N \) of depth \( N=3 \) for a given stream \( X \in \mathbb{R}^{n \times f \times c} \), given that \( n \) is the number of cameras (\( n=906 \)), \( f \) is the number of file increments that make an hour of traffic modes (\( f=11 \)) and \( c \) is the number of channels for traffic modal flows and their stationary status (\( c=13 \)). The collection of computed signatures for all cameras for a given hour is invariant to path reparameterization. This provides 1) a natural characteristic of linear functionals, which only capture the main aspects of the provided path by mapping the sequence of the stream's information rather than mapping the exact position of the path at each occurrence, and 2) the ability to retrieve the original stream of road users and their behaviour from the lower-dimensional signature, minimising computational and memory footprint (Fig. \ref{fig:fig2}-C).

\begin{figure*}[h]
\centering
\includegraphics[width=1\linewidth]{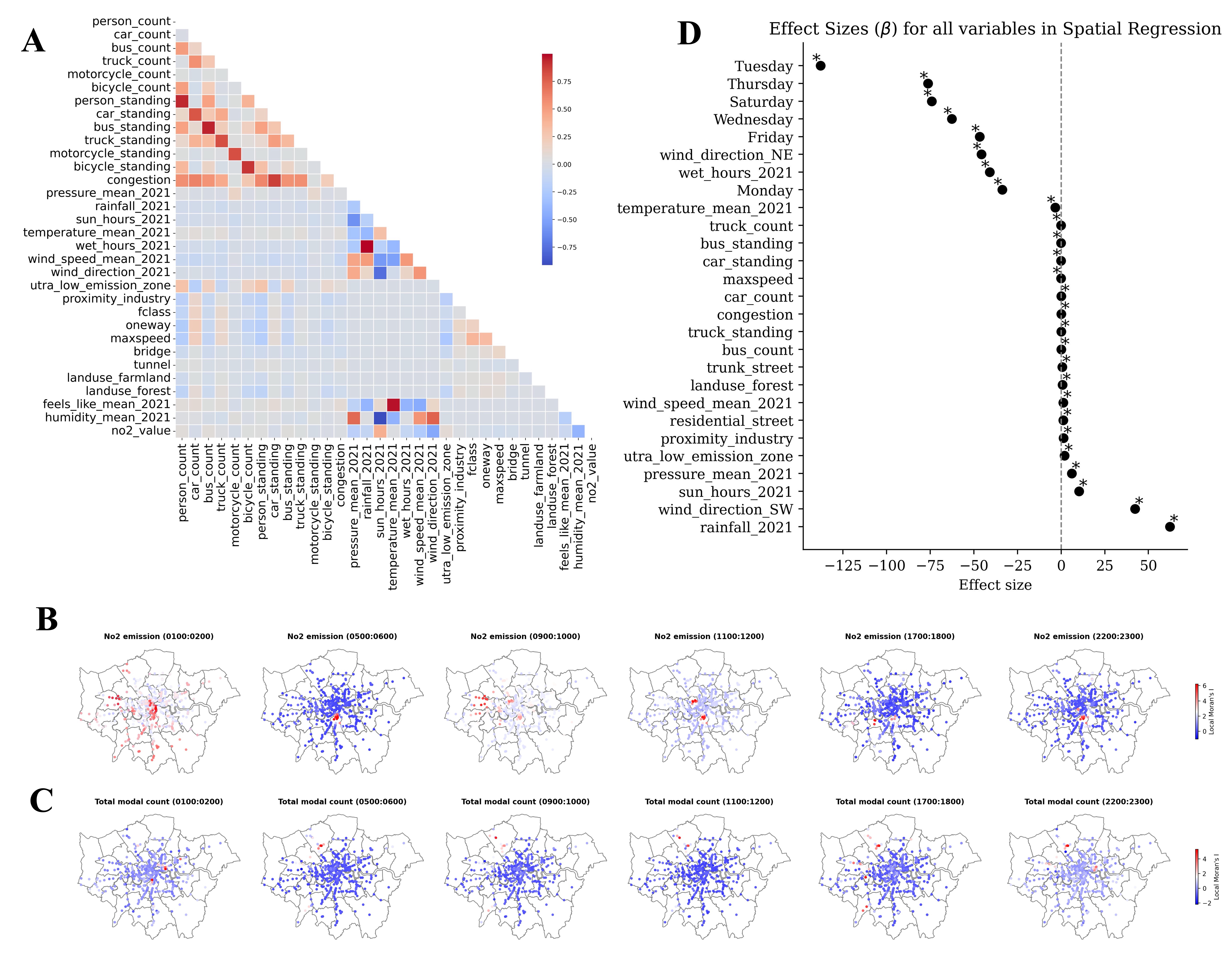}

\caption{The Spatial patterns of NO\textsubscript{2} and traffic at city-scale.  \textbf{(A)} The association between the studied variables relies on Pearson’s correlation. \textbf{(B)} Hot spot analysis using significant Moran’s I z-value (P < 0.05) to highlight the outliers of NO\textsubscript{2} across different hours of the day (the rest of the 24 hours are presented in supplementary). \textbf{(C)} Hot spot analysis using significant Moran’s I z-value to highlight the outliers of total flow across different hours of the day. \textbf{(D)} Statistically significant results ($p<0.05$,$r^2=0.4$, $spatial r^2=0.23$,and $df=88020$) of the spatial two-stage least-square model, variables are shown based on the sign and weight of their $\beta$ value. }
\label{fig:fig3}
\end{figure*}

\begin{figure*}[!t]
\centering
\includegraphics[width=1\linewidth]{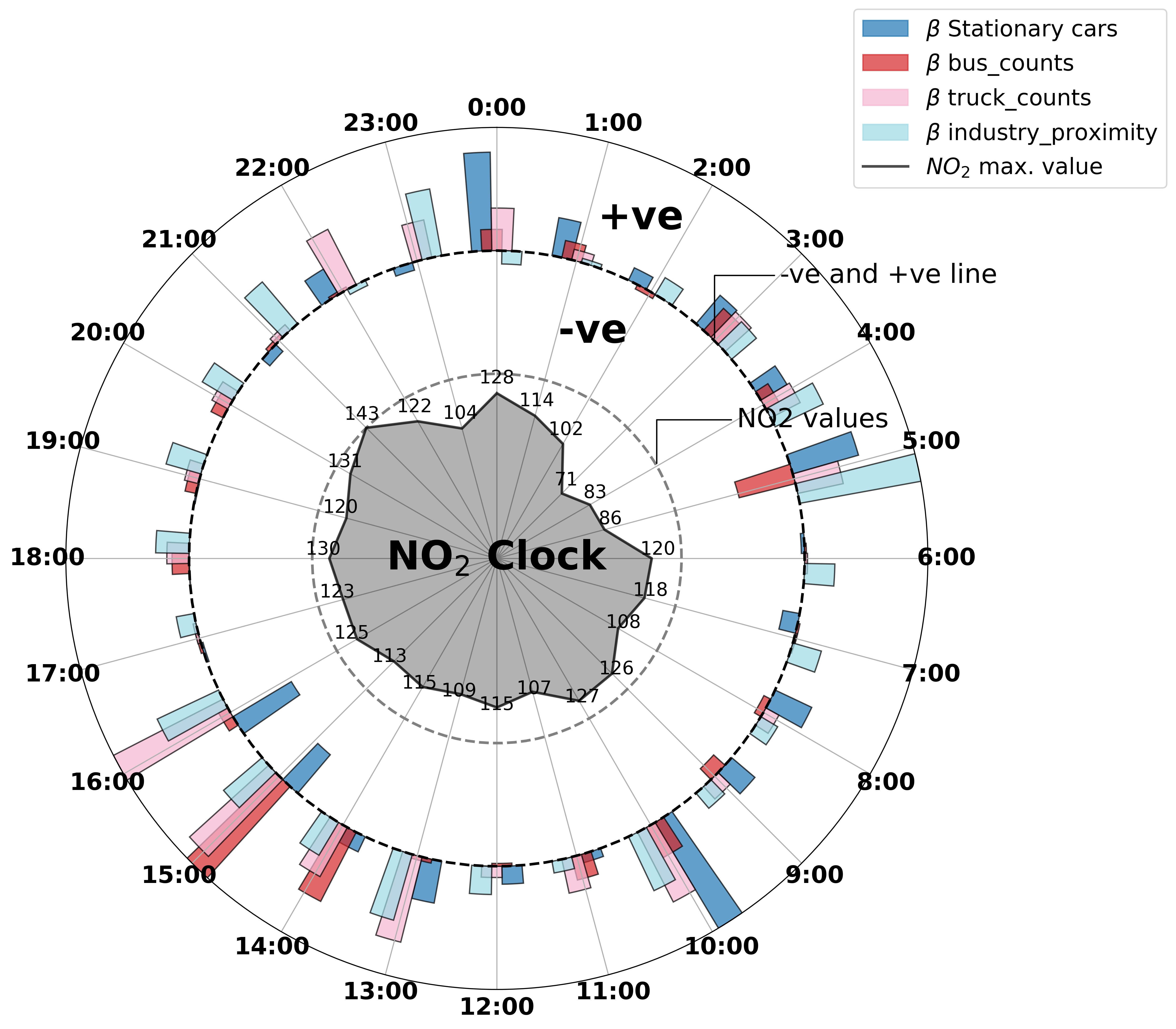}
\caption{NO$_2$ Clock. It shows the hourly average levels of NO$_2$ and the factors influencing these levels, based on a spatial regression model for each hour. It features three concentric circles: the innermost represents the average NO$_2$ concentration per hour, the middle circle shows factors negatively correlated with NO$_2$, and the outermost highlights positively correlated factors. Each factor’s influence is quantified by a $\beta$ value, indicating its effect size relative to the hourly covariates, factors, and overall impact on NO$_2$. All $\beta$ values are standardised across all hours. For simplicity and clarity, the figure displays only four variables, although the full model considers a more extensive range of variables detailed in Table S1.}
\label{fig:no2_clock}
\end{figure*}

\subsection{The effect of location and environment on ground-level NO$_2$}
Geographical factors, such as the proximity to farmland, industrial zones, or various land uses, significantly influence traffic patterns and, as a result, levels of NO\textsubscript{2} (See Fig. \ref{fig:fig3}-A). To investigate the spatial relationship between NO\textsubscript{2} and traffic, we developed a hot spot analysis to cluster total traffic and NO\textsubscript{2} levels based on the spatial dependency of neighbouring high or low values, yielding statistically significant clusters (\(p<0.05\)) of spatial outliers. Here, we show a spatial lag when examining the locations of hot spots for both variables at a given time (See Fig. \ref{fig:fig3}-B and \ref{fig:fig3}-C).

We observe a spatial lag which could be attributed to confounding variables related to environmental factors such as rainfall, wind speed, and direction that either concentrate or disperse emissions from their sources. Moreover, the observed spatial lag may also be linked to the lifetime of NO\textsubscript{2} \cite{source24,lifetime1,lifetime2,lifetime3,lifetime4,lifetime5,lifetime6}, which introduces a temporal delay between traffic emissions and the resultant ground-level concentration of NO\textsubscript{2} detected in a specific area. We will further investigate this in the following section by relying on Granger Causality analysis, which helps in understanding and measuring the delayed effects of traffic emissions on NO\textsubscript{2} ground-level concentrations. However, as a first step, we used a spatial two-stage least squares model to investigate various variables related to geographical characteristics, environment, and day of the week (See Fig. \ref{fig:fig3}-D). We discovered that proximity to industrial zones within one mile (\(\beta=2.156, p=0.000\)), boroughs within Ultra Low Emission Zones (ULEZ) \cite{source7} (\(\beta=3.075, p=0.000\)), wind speed (\(\beta=2.843, p=0.000\)), sun hours (\(\beta=6.438, p=0.000\)), rainfall (\(\beta=43.571, p=0.000\)), South West winds (\(\beta=6.761, p=0.0001\)), congestion (\(\beta=0.060, p=0.000\)), and the change in atmospheric pressure (\(\beta=3.243, p=0.000\)) are more likely to contribute linearly to the level of NO\(_2\) at a given location. Conversely, the number of wet hours in a given day (\(\beta=-33.782, p=0.000\)), the change in average temperature (\(\beta=-2.486, p=0.000\)), North East wind (\(\beta=-26.877, p=0.000\)), average speed limit of a given road (\(\beta=-0.042, p=0.000\)) and proximity to farmland (\(\beta=-0.805, p=0.0013\)) are negatively linear with the emission. We further investigate the temporal dependency of traffic modes within a given hour of the day.

\subsection{The effect of time and the dynamics of traffic modes on ground-level NO$_2$}

Given the relationship between NO\textsubscript{2} levels and total traffic is nonlinear at all times and locations (See Fig. S2-A in supplementary), modelling NO\textsubscript{2} ground-levels requires considering the entire urban landscape as an integrated dynamic system. This approach is especially pertinent because air pollution tends to diffuse and is influenced by numerous factors, such as wind speed, direction, existing green spaces, and proximity to industrial zones or farmlands, in which we have studied. These elements collectively contribute to a nonlinear impact on localised NO\textsubscript{2} levels within the network.

Moreover, NO\textsubscript{2}'s behaviour in the atmosphere adds another layer of complexity to this topic. NO\textsubscript{2} can have variable lifetimes in the air, ranging from a few hours to a whole day depending on meteorological conditions and the presence of other chemical species \cite{source24,lifetime1,lifetime2,lifetime3,lifetime4,lifetime5,lifetime6}. During daylight hours, UV light from the sun can drive photolytic reactions that convert other nitrogen oxides such as NO into NO\textsubscript{2}, further altering the dynamics of air quality. This chemical interplay indicates that emissions and concentrations of NO\textsubscript{2} are fluid, changing not just with traffic flow and industrial activity, but also with the shifting patterns of sunlight and weather.

Despite the complicated dynamics influenced by environmental and chemical processes, there is a discernible linear relationship between NO\textsubscript{2} and types of traffic observed over the course of a day at specific camera locations. This linearity in smaller, more controlled environments suggests that while broader city-wide models must account for complex inter-dependencies and nonlinear behaviours, localised predictions and assessments can successfully utilise simpler linear models. This dichotomy highlights the need for a layered approach in environmental monitoring and management, blending both detailed, location-specific data and broader, systemic perspectives to form a comprehensive understanding of urban air quality.

Building on this, the temporal dynamics play a crucial role in analysing the patterns of NO\textsubscript{2}. To dissect how each factor influences NO\textsubscript{2} levels at distinct times, we implemented two distinct statistical methodologies. Firstly, we employed a spatial regression model for each hour of the day, resulting in 24 unique models. This method helps identify the direct impact of various factors on NO\textsubscript{2} levels at specific hours. Secondly, to explore how each factor may influence future levels of NO\textsubscript{2}, we developed a Granger Causality analysis model for each factor (8 models in total). This technique is particularly useful for pinpointing significant temporal lags and understanding the predictive relationship between the factors and subsequent NO\textsubscript{2} concentrations. These approaches allow us to identify not only the immediate effects of factors on NO\textsubscript{2} levels but also their delayed impacts, thus providing a more comprehensive understanding of the temporal dynamics at play. This layered analysis ensures a more nuanced insight into the cyclic and predictive behaviours of NO\textsubscript{2} in relation to traffic and environmental influences.

Furthering our understanding of the temporal dynamics, Fig. \ref{fig:no2_clock} shows a novel visual representation of the NO\textsubscript{2} clock, showcasing statistically significant linear relations between certain factors and NO\textsubscript{2} levels, characterised for each hour of the day. This graphical display helps to encapsulate NO\textsubscript{2} levels and the main associations observed: for instance, trucks exhibit a consistent linear correlation with NO\textsubscript{2} during midday, night, and the early hours of the morning. In contrast, buses tend to influence NO\textsubscript{2} levels predominantly during the morning and afternoon peak traffic periods. Stationary cars contribute to air pollution during the peak morning hours around 10 am, and their influence extends into midday, primarily while idling in traffic jams. This is different from other periods when stationary vehicles, mainly parked, have little or no impact on pollution. During busy traffic, however, the idling of these cars significantly elevates NO\textsubscript{2} levels. Expanding on these observations, the data also reveals that stationary buses notably contribute to NO\textsubscript{2} during the morning rush hours (8-9 AM). Furthermore, locality factors such as proximity to industrial areas (within a one-mile radius) demonstrate a substantial effect on NO\textsubscript{2} concentrations during specific times—specifically in the evening (7-8 PM) and early morning hours. These insights underscore not only the diverse temporal relationships between different vehicles and NO\textsubscript{2} concentrations but also illuminate the role of geographic and stationary factors in influencing air quality at different times of the day. This level of detail enriches our understanding of urban air pollution dynamics and highlights the critical interplay between temporal, vehicular, and locational determinants in shaping urban NO\textsubscript{2} levels.

Expanding on the analysis of significant temporal lags where specific traffic modes influence and Granger-cause future NO\textsubscript{2} levels, our data demonstrates that the time series of each traffic mode Granger-causes the series of NO\textsubscript{2} with notable statistically significant lagged values. For instance, car flows are likely to Granger-cause NO\textsubscript{2} levels with lag times ranging from 2 to 6 hours, varying by location. Meanwhile, stationary cars manifest a more immediate impact on NO\textsubscript{2} concentrations, typically with a 2-hour lag. In terms of heavier traffic elements, congested traffic flows and stationary buses exert a more prolonged effect on NO\textsubscript{2} levels, showing significant impacts at lags of 5 and 6 hours. Stationary trucks, on the other hand, show a swift influence with only a one-hour lag, suggesting their emissions rapidly integrate into the local atmosphere. Conversely, moving trucks have a more extended influence, where the current flows can predict NO\textsubscript{2} levels up to 5 hours into the future. These findings are also linked to the chemical behaviour of NO\textsubscript{2} in urban air. The timeline of influence observed ties back to the variable atmospheric lifetime of NO\textsubscript{2} \cite{source24,lifetime1,lifetime2,lifetime3,lifetime4,lifetime5,lifetime6}, which can differ from several hours to a full day, influenced by ambient conditions such as sunlight and temperature. Solar radiation promotes the photolytic cycle that converts NO to NO\textsubscript{2}, fundamentally affecting how quickly emissions from traffic transform into atmospheric NO\textsubscript{2}. Therefore, the timing of traffic flows and their characteristic effects on NO\textsubscript{2} can directly correlate with these natural diurnal variations, reinforcing the need to consider both chemical kinetics and traffic dynamics when analysing urban air quality patterns. This multi-faceted approach provides a richer, more accurate depiction of NO\textsubscript{2} ground-level, particularly in dense urban environments where traffic and industrial emissions often overlap.

\begin{figure*}[!t]
\centering
\includegraphics[width=1\linewidth]{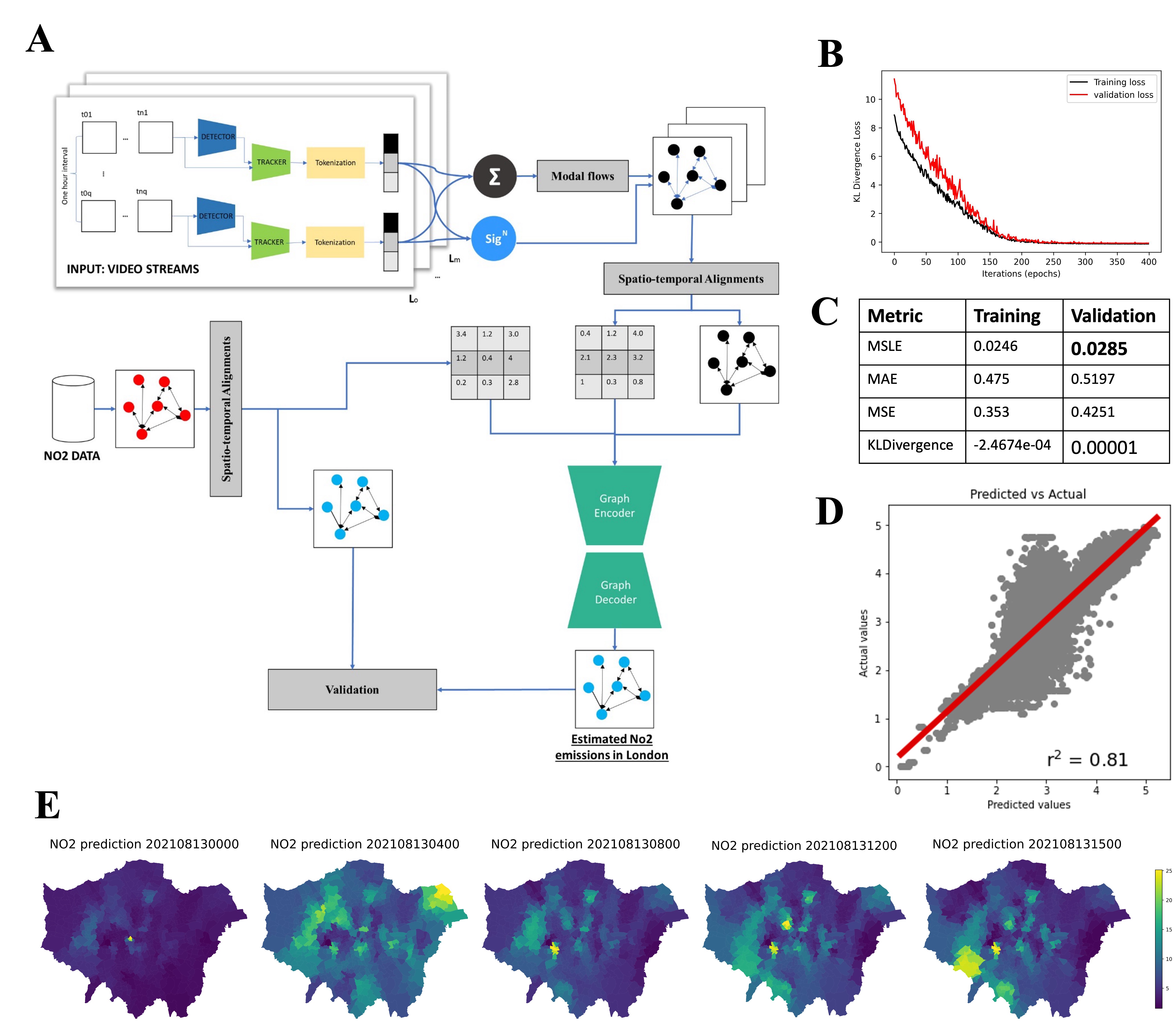}
\caption{Graph-to-Graph model to predict NO\textsubscript{2} surface at a given time from camera feeds. \textbf{(A)} The overall method for the developed a signature-based graph neural network to generate a surface of NO\textsubscript{2} ground-level from camera inputs. The arrows represent the flow of information from the CCTV footage to the prediction of the NO\textsubscript{2} ground-level. \textbf{(B)} A scatter plot for the actual and predicted data for all sensor locations and all dates. \textbf{(C)} The results of training and validation loss and evaluation metrics for training and validation sets. \textbf{(D)} A scatter plot for the actual and predicted data for all sensor locations and all dates. 
\textbf{(E)} NO\textsubscript{2} prediction for different hours of the day, aggregated at a borough level. This figure is created by the first author using python programming. }
\label{fig:graph_model}
\end{figure*}

\subsection{The impact of policies on the dynamics of ground-level NO$_2$}
Not only do factors connected to place and time have a significant impact on NO\textsubscript{2} levels, but so do the measures and regulations implemented in London driven by specific location and time. According to our Granger analysis, the effect of traffic in a given location on the level of NO\textsubscript{2} can appear after several hours,  we found that limiting certain traffic modes, such as trucks, under certain policies (i.e. London Lorry control scheme35) may not be an effective measure for controlling NO\textsubscript{2}, especially in residential areas, given that if the traffic of heavy lorries and trucks is concentrated at night times, its effect will still appear in the morning peak hours when the majority of people are travelling.

Finally, there is still less than one percent of electric cars in London compared to petrol cars, implying that their positive effect on reducing NO\textsubscript{2} levels is likely to be negligible when compared to the entire number of existing petrol and diesel automobile flows. Furthermore, there are still a small number of electric trucks and buses, which we believe, along with stronger steps to restrict emissions from industrial zones, are more likely to cut NO\textsubscript{2} levels in London.

\subsection{Transforming CCTV cameras into NO\textsubscript{2} sensors with a Graph-to-Graph Neural Network}

Building on our understanding of the complex spatiotemporal dynamics of NO\textsubscript{2} levels, we are faced with the challenge of deducing these levels from the complex and nonlinear interactions among various variables. To address this, we developed a Graph-to-Graph deep model using deep learning \cite{source36,source37}, specifically geometric deep learning \cite{source38,source39,source40,source41,source42}, to learn the presented spatiotemporal links and other latent ones that could contribute to the level of NO\textsubscript{2} at a given location while accounting for the dynamics of the entire network, traffic flows in London, and fluid dynamics derived from wind direction and speed. Fig. \ref{fig:graph_model}-A shows the overall conceptual framework of the developed pipeline to forecast NO\(_2\) in London using hourly traffic modal flows in London. The introduced framework also integrates additional secondary data such as weather conditions and spatial features, among other variables (See Fig. S1 in supplementary). The developed model learns in semi-supervised settings from both the states of a given node represented in terms of traffic flows for each mode and the links between nodes represented in their adjacency and their potential influence elsewhere.

Given that the positions of both cameras and environmental sensors are not constrained to one another (as previously shown in Fig. \ref{fig:fig1}-B), the stated problem shifts from identifying regressor values on the same graph to generating a whole graph of a different adjacency matrix than the one given as an input. It is important to note that we used a weighted graph in which fewer links for traffic modes are identified based on the number of nearest neighbours to mimic the actual spatial network, whereas, for the graph of environmental sensors, we used a fully-connected network because air can diffuse freely from one location to another without the spatial constraints of a given network. The model was able to learn to create spatially distributed NO\(_2\) values, resulting in a surface of NO\(_2\) concentration over London at a given hour, using the described method (See Fig. \ref{fig:graph_model}). We also trained several models to assess our method (refer to the methodology section and Table S5).

\section{Discussion}
Monitoring the dynamics of the environment and tracking the progress of environmental policies remains a difficult but critical issue in achieving urban sustainability. In this study, we demonstrated how CCTV cameras and autonomous vision systems using artificial intelligence can aid in monitoring NO\textsubscript{2} levels and evaluating our daily activities in cities that are substantially linked to different NO\textsubscript{2} levels.  We demonstrated how human behaviours related to urban mobility and choice of mobility mode can influence the level of NO\textsubscript{2} differently depending on the dynamics of location and time. We presented novel analyses and insights into the multifaceted nature of the stated issue, such as the impact of time, location, natural and built environments, and urban policies. We demonstrated how CCTV cameras and additional spatial data can be utilised to infer NO\textsubscript{2} levels at the city scale when environmental sensors are unavailable or have sparse coverage when they exit. This technology could benefit numerous cities around the world that lack the infrastructure to monitor pollutants.

Based on this research, various learning lessons and policy implications can be applied to London and other cities across the globe. When it comes to decreasing emissions in cities, the majority of urban policies rely on 1) locational restrictions, 2) temporal constraints, or 3) a combination of temporal and locational constraints. Our findings suggest an alternative approach for developing environmental legislation that considers overall emissions across all locations and times of day. We demonstrated that there are temporal lags between current traffic and their impact on future NO\textsubscript{2} emissions. This implies the need for new policy reform that considers a minimal overall emission during different hours of the day rather than temporal constraints and concentrating unwanted traffic at a given time of the day. Given that our findings suggest that if trucks, for example, only drive at night within the inner parts of the city, their impact on emissions will be more likely to appear in the morning (with a lag of up to 6 hours), where more people may be affected.

\subsection{Limitations}

There are still data uncertainties in big data, particularly video streams, making the presented traffic counts an approximation of day-to-day operations in Greater London. These uncertainties stem from factors such as camera field of view, obstruction, or biases due to the chosen locations for sensors \cite{source43}. Effective sensor deployment is essential for urban-scale monitoring to ensure comprehensive coverage and reliable data collection \cite{song2022toward}; however, this study assumes both cameras and NO\textsubscript{2} sensors are provided and does not cover sensor placement. The placement of CCTV cameras can introduce biases into our NO\textsubscript{2} predictions. Cameras are typically located in high-traffic areas, which may not fully represent overall urban air quality. We have discussed this limitation and the measures taken to mitigate its impact. As a result, we considered numerous strategies such as recognising outliers and data stationary wherever it is acceptable for a certain method. Furthermore, many features derived from data tend to follow rational thinking of patterns that are predicted to be shown, according to descriptive analysis. For example, cars contribute to traffic congestion but not bicycles, the two traffic peaks of a given day when the total flow is distributed throughout all hours of a given day, and the negative relationship between cycling and the level of NO\textsubscript{2}, among other things.

While the presented models require minimal inference time ($<$0.1sec) to generate NO\textsubscript{2} at a given hour, it is critical to understand the centralised computational requirements for computing and extracting traffic flow data from CCTV video feeds at scale. The supplied data across all cameras and all days were retrieved using 84 days of computing on a single GPU. Accordingly, finding alternative solutions to minimise the time for deployment at a scale of a given city is necessary. Two approaches can be used to do this: 1) learning the complete traffic flow at a city level for a given time from only fewer camera inputs, and 2) decentralised computations at the edge by relying on AI-enabled cameras that deploy lightweight models on minimal hardware sensors. This method might enable real-time NO\textsubscript{2} data processing and inference, as well as proactive sensing of its determinants at any given time and place.


\section{Methods}

Our study enhances CCTV-based analysis and NO\textsubscript{2} monitoring by demonstrating the use of existing infrastructure for environmental sensing, which is especially beneficial for cities with limited access to specialised air quality sensors. Our method can be implemented in other cities with a sufficient number of CCTV cameras. For city-wide NO\textsubscript{2} prediction, our model utilises traffic data extracted from cameras, along with environmental and locational factors, and the computed signature of this data to predict city-wide NO\textsubscript{2} levels. The camera and NO\textsubscript{2} sensor locations do not have to coincide, providing flexibility in applying and transferring this method to any location. The input data comprises traffic data extracted from CCTV camera footage, including various road users' modal flows and their stationary statuses. Additionally, we included environmental factors such as average wind speed, wind direction, wet hours, sun hours, rainfall, average pressure, average humidity, average temperature, and proximity to industrial zones. The ground truth data for training and validating our models were sourced from hourly NO\textsubscript{2} sensor measurements across multiple locations within London. The target features for our models were the NO\textsubscript{2} levels, either at specific sensor locations or across a generated surface for city-wide prediction. By integrating the computed signature of the traffic data with locational and environmental features, our models provided accurate predictions of NO\textsubscript{2} levels, demonstrating the feasibility of using existing CCTV infrastructure for environmental monitoring and policy-making. Here, we describe the materials and methods utilised to develop this research.

Here we describe our materials and the different methods utilised to develop this research.

\subsection*{Materials}

All raw data sources can be accessed online.

\begin{enumerate}
    \item \textbf{London CCTV data:} We collected video streams that represent 892 unique camera locations across London for 56 different hours of scattered days in the year 2021. This data includes 65,493,858 sequential frames, in which the total data or a subset of it has been used for different analyses represented in the paper.  We also collected additional video data for a given camera (ID) for a given hour (12 am-1 pm) across all the days of the year to show the seasonal dynamics of traffic patterns.  The data can be accessed via API permissions from Transport for London (TfL).

    \item \textbf{Hourly NO\textsubscript{2} data:} We extracted hourly N02 data of 144 unique sensors that link to the extracted video hours. The raw data can be accessed through an API from London Air: 
    \url{https://www.londonair.org.uk/london/asp/annualmaps.asp}

    \item \textbf{Weather data:} We linked the camera and NO\textsubscript{2} data to the weather day based on a day resolution. We included nine variables as a representation of the environmental conditions of a given day. This data, includes 1) average wind speed, 2) wind direction, 3) wet hours, 4) sun hours, 5) rainfalls, 6) average pressure, 7) average humidity, 8) average temperature, and  9) average feels like temperature. The raw data can be accessed from: 
 \url{http://nw3weather.co.uk/wxdataday.php?vartype=wmean&year=2021}

    \item \textbf{Spatial data:} We used GIS shapefile data for the spatial representations of London’s boroughs, spatial network, and the boundary of the city. The spatial network data included 1) whether a given street is two-directional, 2) average speed and 3) the type of the street. The raw data can be accessed from Greater London Authority: \url{https://data.london.gov.uk/dataset/statistical-gis-boundary-files-london}

    \item \textbf{Car flows based on engine types:} To evaluate the percentage of electric cars to petrol and diesel ones in each borough, we used the traffic flow data provided by London Council. This data is used for statistical analysis to account for the ratio of cars based on the engine types that we observe in CCTV cameras at a given location. The data is entitled: “laei-2019-major-roads-vkm-flows-speeds” and can be accessed from: \url{https://data.london.gov.uk/dataset/london-atmospheric-emissions-inventory--laei--2019}

    \item \textbf{Proximity to industrial zones:} We used Strategic Industrial Location Points to calculate a buffer zone of 1 mile and account for the camera’s locations that are within this zone. The raw dataset can be accessed online from:
\url{https://data.london.gov.uk/dataset/strategic-industrial-location-points-london-plan-consultation-2009}

\end{enumerate}

\subsection*{Extracting road users from video streams }
To extract the six types of road users from video streams and their relevant information, we used a deep learning framework that comprises multiple deep models including, You Look Only Once (YOLO) architecture \cite{source44,source45}. Particularly, we relied on  YoloV5m \cite{source46} coupled with DeepSort architecture \cite{source47} to detect and track road users throughout a given video file. DeepSort architecture is built on a deep learning model with Sort algorithms \cite{source48} to account for object occlusion. We used a pre-trained weight of YOLOV5m model trained on COCO dataset \cite{source49}. It's worth mentioning that computing this data and transforming it from raw video streams to vector data took almost 18 hours for analysing one hour across all cameras for a given day (84 days in total) on a single GPU.

\subsection*{Projecting road users in a bird’s eye view map}
Transforming moving objects from CCTV footage to a top-view perspective is crucial for accurately analysing and verifying various traffic factors. This perspective allows for the consistent identification and tracking of road users, regardless of obstructions within the camera's field of view. By projecting the traffic data onto a bird’s eye view, we can effectively distinguish between stationary and non-stationary road users, offering a clearer and more precise understanding of traffic dynamics. Additionally, this transformation ensures geographic consistency when integrating data with mapping services, enhancing the overall spatial accuracy of our traffic flow analyses. This step is integral to mitigating common issues associated with perspective distortion in street-level imagery, ensuring reliable data for predicting NO\textsubscript{2} levels.

We relied on the TopView framework to transform objects from the camera view to the bird’s eye view without knowing the camera models that include both intrinsic and extrinsic parameters \cite{source53}. The framework relies on a deep learning model to detect the vanishing point (VP) in a given scene, whereas four points in the camera view can be automated and correspond to four points in world coordinates and accordingly to a bird’s eye view map based on geometric transformation and homography\cite{source50,source51,source52,source53}. We used the VP model and paired points in the two views to determine the homography matrix \( H \) as follows:

\begin{align}
\begin{bmatrix}
z_i x_i' \\
z_i y_i' \\
z_i
\end{bmatrix} &= H \begin{bmatrix}
x_i \\
y_i \\
1
\end{bmatrix}, \\
\text{where} \quad \text{dst}(i) &= (x_i', y_i'), \quad \text{src}(i) = (x_i, y_i), \quad i=0,1,2,3 \nonumber
\end{align}

Given that src and dst are the coordinates of the quadrangle vertices in the camera view and world coordinates respectively, \((x_i, y_i)\) and \((x_i', y_i')\) are the paired coordinate points in the camera and the bird’s eye view planes respectively and \( H \) is the transformation of the homography matrix that is computed as:

\begin{equation}
H = \begin{bmatrix}
h_{00} & h_{01} & h_{02} \\
h_{10} & h_{11} & h_{12} \\
h_{20} & h_{21} & h_{22}
\end{bmatrix}
\end{equation}

Given that \( H \) is calibrated based on the four paired points that are produced by the camera and top-view planes, respectively. And therefore, the detected object in the camera plane may be changed into the top-view plane by resolving \( H \). For further explanation, see the full explanation of the TopView method\cite{source54}.

\subsection*{Tokenizing road users and counting flows}
To detect modal flows, we first tracked road users in a given file, where each road user has a unique ID, and then the number of road users is counted throughout the file. The road users are vectorized based on their tracked ID data and visualised based on when they appear and disappear in the video files while keeping in mind that multi-dimensional data, such as stationary status, road user categories, and trajectory line in the bird’s eye view, has been retrieved. 

\subsection*{Ranks of traffic composition}
We estimated the ranks of traffic composition by separating the total counts into unique values that indicate nodes (n=1,n=2, etc.) to grasp the collective behaviour of road users from the local site of all cameras to the city scale. Following that, we computed the unique patterns across each node value (i.e., in the case of n=2, the possible scenarios are vehicle and person, car and car, etc.) and assigned a unique id to each unique pattern. Instead of summing the counts for each mode, we sum the structure at the city level, for example (1-1 + 2-2 + 3-1) up to the number of files.

\subsection*{Granger Causality}
Granger causality \cite{source55,source56,source57,source58} is tested in the context of linear regression, and it is significant when the previous values of a given variable \( X_1 \) contribute to the forecasting of the current value of variable \( X_2 \) or vice versa. By considering a bivariate autoregressive model for these two variables:

\begin{equation}
X_1(t) = \sum_{j=1}^{p} A_{11,j} X_1(t-j) + \sum_{j=1}^{p} A_{12,j} X_2(t-j) + \varepsilon_1(t)
\end{equation}

\begin{equation}
X_2(t) = \sum_{j=1}^{p} A_{21,j} X_1(t-j) + \sum_{j=1}^{p} A_{22,j} X_2(t-j) + \varepsilon_2(t)
\end{equation}

given that \( p \) represents the number of lagged observations in the model order. The matrix \( A \) comprises the coefficients of the model such as the contributions of each lagged observation to the predicted values of \( X_1(t) \) and \( X_2(t) \), and \( \varepsilon_1 \) and \( \varepsilon_2 \) are the model residuals for each time series.

If the coefficients in \( A_{12} \) are all considerably different from zero, then \( X_2(t) \) Granger causes \( X_1(t) \). The model significance is tested by computing an F-test of the null hypothesis that \( A_{12} = 0 \), assuming that the stationarity of the covariance on \( X_1(t) \) and \( X_2(t) \). The logarithm of the associated F-statistic can be used to determine the size of a Granger causality interaction \cite{source59,source60}.

According to the Granger test, it is worth mentioning that causality is evaluated on the grounds that 1) the cause precedes the effect and 2) the cause has specific knowledge about the potential outcomes of its impact. To demonstrate the significant findings of Granger testing, we show the results of four parameters, including the parameters for the F-test and ssr-F-test which are based on the F-distribution and the parameters for the ssr-based chi-squared test and the likelihood ratio test, which are based on the chi-square distribution.

\subsection*{Spatial weight}

Using the K-Nearest Neighbour weights technique \cite{source61}, we estimated the spatial weight matrix $(\omega_{ij_{t}})$ between the various camera sites at a particular time ( t ). It is a set of neighbours defined by distance-based weights based on ( K ) observations. We investigated several ( K ) values and found that 10 was the best approximation of the number of neighbours where the different camera locations closely matched the actual spatial network. We computed a dynamic spatial weight that differs based on the point representation of a given time. We utilised the estimated spatial weight in many analyses, including spatial clustering, the spatial regression model, and the Graph model.

\subsection*{Spatial clustering and outliers detection}

We computed statistically significant spatial clusters and hot-spot analysis based on Local Moran’s I \cite{source62,source63}. If the value of \( I \) is positive, it means that a feature is part of a cluster and that it is surrounded by other features that have similar attributes that are either high or low. A negative value for \( I \) implies that an outlier feature has nearby features with values that differ from its own. For the cluster or outlier to be regarded as statistically significant, the \( p \)-value for the feature must be low enough in both cases.

\begin{equation}
I_i = \frac{(x_i - \bar{X})}{S_i^2} \sum_{j=1, j \neq i}^{n} \omega_{ij} (x_j - \bar{X})
\end{equation}

Given that \( x_i \) is the attribute for feature \( i \), \( \bar{X} \) is the mean for the corresponding attribute, \( \omega_{ij} \) is the spatial weight between feature \( i \) and \( j \).

\begin{equation}
S_i^2 = \frac{\sum_{j=1, j \neq i}^{n} (x_j - \bar{X})^2}{n - 1}
\end{equation}

Given that \( n \) is the total number of features.

The Z-score for the statistics is defined as:

\begin{equation}
Z_{I_i} = \frac{I_i - E[I_i]}{\sqrt{V[I_i]}}
\end{equation}

\begin{equation}
E[I_i] = - \frac{\sum_{j=1, j \neq i}^{n} \omega_{ij}}{n - 1}
\end{equation}

\begin{equation}
V[I_i] = E[I_i^2] - E[I_i]^2
\end{equation}

\subsection*{Spatial Regression model }
Given the geographical dependency of the observed variables, we employed a spatial regression model \cite{source64,source65} rather than a simple regression model to assess the statistically significant links between NO\(_2\) levels and the various values of road users and the built environment. We explored three different approaches in which spatial weight can be applied including, the spatial dependency model, spatial error model, and spatial lag model. First, in the spatial dependency model, the previously computed spatial weight \( \omega_{ij} \) is accounted in the model as an additional independent variable as follows: 

\begin{equation}
\log(P_i) = \alpha + X\beta + WX\gamma + \varepsilon
\end{equation}

\begin{equation}
\log(P_i) = \alpha + \sum_{k=1}^{p} X_{ij} \beta_j + \sum_{k=1}^{p} \left( \sum_{j=1}^{N} \omega_{ij} x_{jk} \right) \gamma_k + \varepsilon_i
\end{equation}

Second, in the spatial error model, we account for the spatial dependence in the model residual as follows: 

\begin{equation}
\log(P_i) = \alpha + \sum_{k=1}^{p} X_{ki} \beta_k + \mu_i
\end{equation}

\begin{equation}
\mu_i = \lambda_{ulag-i} + \varepsilon_i
\end{equation}

\begin{equation}
\lambda_{ulag-i} = \sum_j \omega_{ij} u_j
\end{equation}

Last, the Spatial lag model can be computed as:

\begin{equation}
\log(P_i) = \alpha + \rho \log(P_{lag-i}) + \sum_{k=1}^{p} X_{ki} \beta_k + \varepsilon_i
\end{equation}

\subsection*{NO\textsubscript{2} Surface construction from points}
We also relied on the triangulation method to generate a 3D surface from the sensors' unique locations by creating triangles by specifying their corners based on three given points. 

\subsection*{Signature of paths }
This research is concerned with multi-level temporal scales that go from the temporal representation of a certain sequence of a video file at a given location to the hourly temporal representation of video files that can correspond to the temporal scale of NO\(_2\) Data. As a result, in addition to depending on a straightforward strategy of summing the data increments of a given hour at a specific site, we relied on rough path theory and path signature \cite{source32,source33,source34,source66,imagesig} to summarise the multidimensional temporal representation of the presented data. As a result, we developed a method for summarising the key patterns within the video increments of an hour without losing the raw data relying on signature due to its invariance to reparameterisations. The truncated signature of a path \( \gamma_t \) at a given depth \( N \) at a given hour is defined as:

\begin{equation}
S_{a,b}(\gamma_t) = \bigoplus_{n=0}^{N} S_{a,b}^n(\gamma), \quad \text{given that} \quad S_{a,b}^n(\gamma_t) = \frac{1}{n!} (\gamma_b - \gamma_a)^{\otimes n}
\end{equation}

The signature transform given that \( \text{Sig}^N = S(\mathbb{R}^d) \to \prod_{n=1}^{N} (\mathbb{R}^d)^{\otimes n} \) is computed as:

\begin{align}
\text{Sig}^N(X) = &\left( \int_{0 < t_1 < \ldots < t_n < 1} \frac{df}{dt}(t_1) \otimes \ldots \otimes \frac{df}{dt}(t_n) \, dt_1 \ldots dt_n \right)_{1 \leq n \leq N} \\
&\text{for } 1 \leq n \leq N
\end{align}

The log signature of \( \gamma_t \) is defined as:

\begin{align}
\log S_{a,b}(\gamma_t) = &\bigoplus_{n=0}^{N} \frac{(-1)^{n-1}}{n} \left( \hat{S}_{a,b}^n(\gamma) \right)^{\otimes n}, \\
&\text{given that } S_{a,b}^0(\gamma_t) = 1 \text{ and } \hat{S}_{a,b}(\gamma_t) = \bigoplus_{n=1}^{N} S_{a,b}^n(\gamma_t)
\end{align}

\subsection*{Graph model architectures}

We developed an undirected weighted Graph \(G(V, E, A, \omega)\), where \(V\) is the set of nodes with \(|V| = N\) is the number of nodes, \(E\) represents the set of the edges of the graph, \(A\) is the adjacency matrix and is an \(N \times N\) sparse matrix, and \(\omega_{ij}\) represents the adjacency matrix between node \(v_i\) and \(v_j\). A graph signal \(f: V \to \mathbb{R}\) represents a function defined on the vertices of a graph \(G\) which maps every vertex \({v_i}_{i=1,\ldots,N}\) to a real number \(f_i\). The graph signal \(f\) can be projected to the eigenvectors of the Laplacian matrix \(L\) and by assuming that \(\lambda_l\) and \(\mu_l\) are the \(l_{th}\) eigenvalue and eigenvector of the Laplacian matrix \(L\), the graph Fourier transform \(\hat{f}\) of the graph signal can be defined as:
\begin{equation}
GF[f](\lambda_l) = \hat{f}(\lambda_l) = \langle f, \mu_l \rangle = \sum_{i=1}^{N} f(i) \mu_l^*(i), \quad \text{given that } \mu_l^* = \mu_l^T
\end{equation}
In the context of graph \cite{source37,source39,source41}, the convolution operation between two functions \(f\) and \(g\) can be applied by relying on graph Laplacian eigenvectors and can be defined as:
\begin{equation}
(f * g) = IGF[GF[f] \cdot GF[g]], \quad (f * g)(i) = \sum_{l=0}^{N-1} \hat{f}(\lambda_l) \hat{g}(\lambda_l) \mu_l(i)
\end{equation}
The Graph model comprises \(L^{th}\) graph convolution layers, in which each layer constructs an embedding for each node by fusing the embeddings of the neighbours of a given node from the previous layer as follows:
\begin{equation}
Z^{(l+1)} = A' X^{(l)} W^{(l)}, \quad X^{(l+1)} = \sigma(Z^{(l+1)})
\end{equation}
given that \(X^{(l)} \in \mathbb{R}^{N \times F_l}\) represents the embedding of the \(l\)-th layer for all \(N\) nodes, \(X^{(0)} = X\), \(A'\) is the weighted and normalized adjacency matrix, \(W^{(l)} \in \mathbb{R}^{F_l \times F_{l+1}}\) is the feature transformation matrix that will be learned, and \(\sigma(\cdot)\) is the activation function for which we implemented an element-wise ReLU.

We also used a Graph Attention layer \cite{source38}, given an input of a set of node features \( \mathbf{h} = \{\mathbf{h}_1, \mathbf{h}_2, \ldots, \mathbf{h}_N\} \), \(\mathbf{h}_i \in \mathbb{R}^F\) where \(F\) is the number of features in each node. The layer outputs a new set of node features \( \mathbf{F}', \{\mathbf{h}'_1, \mathbf{h}'_2, \ldots, \mathbf{h}'_N\} \), \(\mathbf{h}'_i \in \mathbb{R}^{F'}\). The linear transformation of the layer is applied to each node, parameterised by a weight matrix, \(W \in \mathbb{R}^{F' \times F}\), in which a shared attentional mechanism is performed on the nodes to indicate the importance of features in a given node \(j\) to node \(i\). Their attention coefficients are defined as: 
\begin{equation}
e_{ij} = a(\mathbf{W}\mathbf{h}_i, \mathbf{W}\mathbf{h}'_j)
\end{equation}
The attention mechanism \(a\) can be defined as a single feedforward layer, parametrized by a weight vector \(\mathbf{a} \in \mathbb{R}^{2F}\), activated by a LeakyReLU nonlinearity, its coefficients can be defined as:
\begin{equation}
\alpha_{ij} = \frac{\exp(\text{LeakyReLU}(\mathbf{a}^T [\mathbf{Wh}_i \| \mathbf{Wh}'_j]))}{\sum_{k \in N_i} \exp(\text{LeakyReLU}(\mathbf{a}^T [\mathbf{Wh}_i \| \mathbf{Wh}'_k]))}
\end{equation}
Given that \(\|\) represents the concatenation operation and \(.T\) represents transposition.

We have experimented with both graph layers, in which we have trained multiple models for two different tasks that take different inputs and generate different outputs, as follows: 

\subsection*{Task 1: Estimating the NO\textsubscript{2} surface for a given hour from traffic flows data}
This model takes an input \(X\) where \(X \in \mathbb{R}^{H \times N \times C}\) and generates NO\textsubscript{2} levels across London for a given hour (\(Y\)) as \(Y \in \mathbb{R}^{H \times M}\), where \(H\) is the number of unique hours, \(N\) is the number of cameras’ locations, \(C\) is the number of features, including modal flows and locational urban features, and \(M\) is the number of NO\textsubscript{2} sensors’ locations where \(M \neq N\).

\subsection*{Task 2: Estimating NO\textsubscript{2} at a given location from the graph knowledge of traffic flows}
This model takes an input \(X\) where \(X \in \mathbb{R}^{H \times N \times C}\) and generates  NO\textsubscript{2} concentration at a given location for a given hour (\(Y\)) as \(Y \in \mathbb{R}^{H}\), where \(H\) is the number of unique hours, \(N\) is the number of cameras’ locations, \(C\) is the number of features, including modal flows and locational urban features.

Further results for all models and their hyperparameters are provided in supplementary, in table 5.

\subsection*{Training Objective Loss}
We trained our models based on Mean Squared Logarithmic Error (MSLE), defined as: 
\begin{equation}
\text{Loss} = (\log(x + 1) - \log(y + 1))^2
\end{equation}
Given that \(x\) and \(y\) are the true and predicted values of NO\textsubscript{2} levels of a given location at a given hour. 

\subsection*{NO\textsubscript{2} Model Validation Metrics}
Furthermore, we computed different metrics to compare the results of the trained models and to validate their performances. We calculated Kullback–Leibler divergence, or known as relative entropy denoted as \(D_{KL}(P \| Q)\) and is defined as: 
\begin{equation}
D_{KL}(P \| Q) = \sum_{x \in X} P(x) \log \left( \frac{P(x)}{Q(x)} \right)
\end{equation}
given that \(P\) and \(Q\) are two discrete probabilities distributions on the same sample space \(X\) representing the distributions of true values and predicted ones. Second, we computed Mean Absolute Error (MAE), known as L1 loss, and is defined as:
\begin{equation}
L_1(x, y) = \frac{\sum_{i=1}^{n} |y_i - x_i|}{n}
\end{equation}
given that \(y_i, x_i\) are the predicted and true values of NO\textsubscript{2} levels respectively, and \(n\) is the batch size.

\subsection*{Training Setup and Implementation Details}
We report on 20 models with different hyperparameters and architecture (See table 5 in supplementary). All models are trained based on the input of the normalized numerical values of traffic flows and categorical values of all factors explained previously after being factorised and transformed into dummies. However, they vary, in terms of input, based on whether 1) the computed signature is included as an input, 2) the adjacency matrix of the NO\textsubscript{2} sensor data is included, besides the adjacency matrix of the CCTV cameras and 3) the number of nearest neighbours when computing the edge or the adjacency matrix. To account for the current state-of-the-art baselines, we trained different architectures as follows:

\textbf{Graph Attention Model:}
We trained several models based on the architecture of three graph attention layers, in which each layer comprises 6 attention heads and each computing 907 features, followed by an ELU nonlinearity layer. The final layer is used to output NO\textsubscript{2} values, containing 1 feature (in case of inferring a NO\textsubscript{2} value for a single location) or N features based on the number of NO\textsubscript{2} sensors (In case of inferring spatially distributed NO\textsubscript{2} values or inferring traffic flows in N cameras), followed by activation of a logistic sigmoid function. We applied dropout \cite{source66,source67} within the three-layer blocks to avoid over-fitness. We trained the models based on a batch size of 8 graphs for 100 training cycles (epochs). All models are initialized by Glorot initialization \cite{source68} and trained to minimise the introduced loss function based on Adam stochastic gradient descent optimiser \cite{source69}, with an initial learning rate of 0.01 and an early stopping strategy based on the validation loss, with patience of 20 epochs.

\textbf{Graph Convolution Model:}
Similar to the graph attention model we trained a graph model based on three graph convolution layers instead of the graph attention model. We followed a close implementation of the originally introduced method and best practice guidelines to provide a baseline \cite{source37}. All models based on graph convolution are trained based on hidden units of 50 features and a dropout of 0.5. The models are trained based on a batch size of 64 using Adam optimiser, with an initial learning rate of 0.01 and an early stopping strategy based on the validation loss, with a patience of 20 epochs.

\textbf{Multi-Branch Graph Model:}
This model architecture takes six inputs, including camera nodes, camera edge, categorical feature, numerical features, and environmental sensor adjacency matrix (or five inputs without signature). Each input is encoded through an isolated branch of three 1D Convolutional layers of 32 filters, kernel size of 1 and activated with a ReLU function followed by a Dropout layer of size 0.4. Finally, a Flatten layer and a fully connected layer of 50 features are used. After each encoder, all outputs are concatenated and passed to a Fully connected layer and a final output of N features that is equivalent to the number of nodes in The NO\textsubscript{2} surface for a given hour, activated based on the Softplus function. The model is trained with a batch size of 2 graphs, and for 300 epochs, following similar procedures of the previous architectures.

\textbf{Transformer Model: }
We also trained several models based on transformer architecture without an explicit graph structure like the case in the first graph architectures. We replaced the convolutional layer in the introduced architecture of the multi-branch graph model, with three transformer layers. Each transformer layer comprises 6 attention heads and projection dimensions of 907 features, followed by a skip connection, a normalization layer, a Multi-layer Perceptron (MLP) and a second skip connection layer. Afterwards, we used a layer normalization and calculated attention weights, in which the product of both attention weights and the previous layer outputs are passed to a single fully connected layer. The final layer is used to output NO\textsubscript{2} values, containing 1 feature (in case of inferring a NO\textsubscript{2} value for a single location) or N features based on the number of NO\textsubscript{2} sensors (In case of inferring spatially-distributed NO\textsubscript{2} values or inferring traffic flows in N cameras), followed by activation of a Softplus function. We also applied dropout to avoid over-fitness. We trained the models based on a batch size of 2 for 300 epochs. We used AdamW stochastic gradient descent optimiser to minimise the introduced loss function, with an initial learning rate of 0.001 and an early stopping strategy based on the validation loss, with a patience of 20 epochs.

\textbf{Evaluating Models Under Different Environmental Conditions: } We trained various model architectures with different hyperparameters to create a baseline and validate our method using different evaluation metrics (see Table S5). We conducted an error analysis to assess model performance under various weather conditions. This involved analysing the impact of factors like rain, wind speed, and temperature on NO\textsubscript{2} levels, providing insights into the robustness of our models. Additionally, we evaluated model performance over different time periods, such as hourly, daily, weekly, and monthly intervals, to ensure consistency. We also assessed the models at different locations within the study area to account for spatial variability in NO\textsubscript{2} levels. Through these thorough evaluations, we aim to demonstrate the reliability and accuracy of our models in predicting NO\textsubscript{2} levels under various real-world conditions. In our study, several models showed promising results. For example, the Graph Convolutional Model with Signature (Model ID 1) exhibited good performance with a mean squared logarithmic error (MSLE) of 0.0375 and a mean absolute error (MAE) of 0.6558. This model integrates graph convolution operations, which are effective in capturing spatial dependencies in the data. The Attention-based Graph Model without Signature (Model ID 3) introduces attention mechanisms within the graph neural network framework. Although this model has significantly more parameters (120,342,324) and longer training time, it presented robust results with an MSLE of 0.0454 and an MAE of 0.6842. The attention mechanism helps in focusing on the most relevant parts of the graph, providing better feature representation. At City wide prediction, the Conv1D-based multiple branch model with Signature (Model ID 19) demonstrated strong performance, providing accurate predictions and showing a high correlation with actual NO\textsubscript{2} levels. By incorporating signature information (N=3), the model enhances its predictive accuracy. The multi-branch design allows the model to process various data aspects in parallel, boosting its learning capacity.

\section*{Acknowledgment}
Mohamed Ibrahim was funded in part by The Alan Turing Institute, the Data Centric Engineering Programme (under the Lloyd’s Register Foundation grant G0095). Terry Lyons was funded in part by the EPSRC [grant number EP/S026347/1], in part by The Alan Turing Institute under the EPSRC grant EP/N510129/1, the Data Centric Engineering Programme (under the Lloyd’s Register Foundation grant G0095), the Defence and Security Programme (funded by the UK Government) and the Office for National Statistics and The Alan Turing Institute (strategic partnership) and in part by the Hong Kong Innovation and Technology Commission (InnoHK Project CIMDA).

\bibliography{sn-bibliography}

\section*{Data availability}
All raw data sources are listed in the Materials and Methods section.

\section*{Supplementary Materials}
\renewcommand{\thefigure}{S\arabic{figure}}  
\setcounter{figure}{0}  

\subsection*{S1 - S2 Figures}
\subsection*{S1 - S5Tables}

\newgeometry{left=0.5in, right=0.5in}  

\begin{figure*}[h]
\centering
\includegraphics[width=1\linewidth]{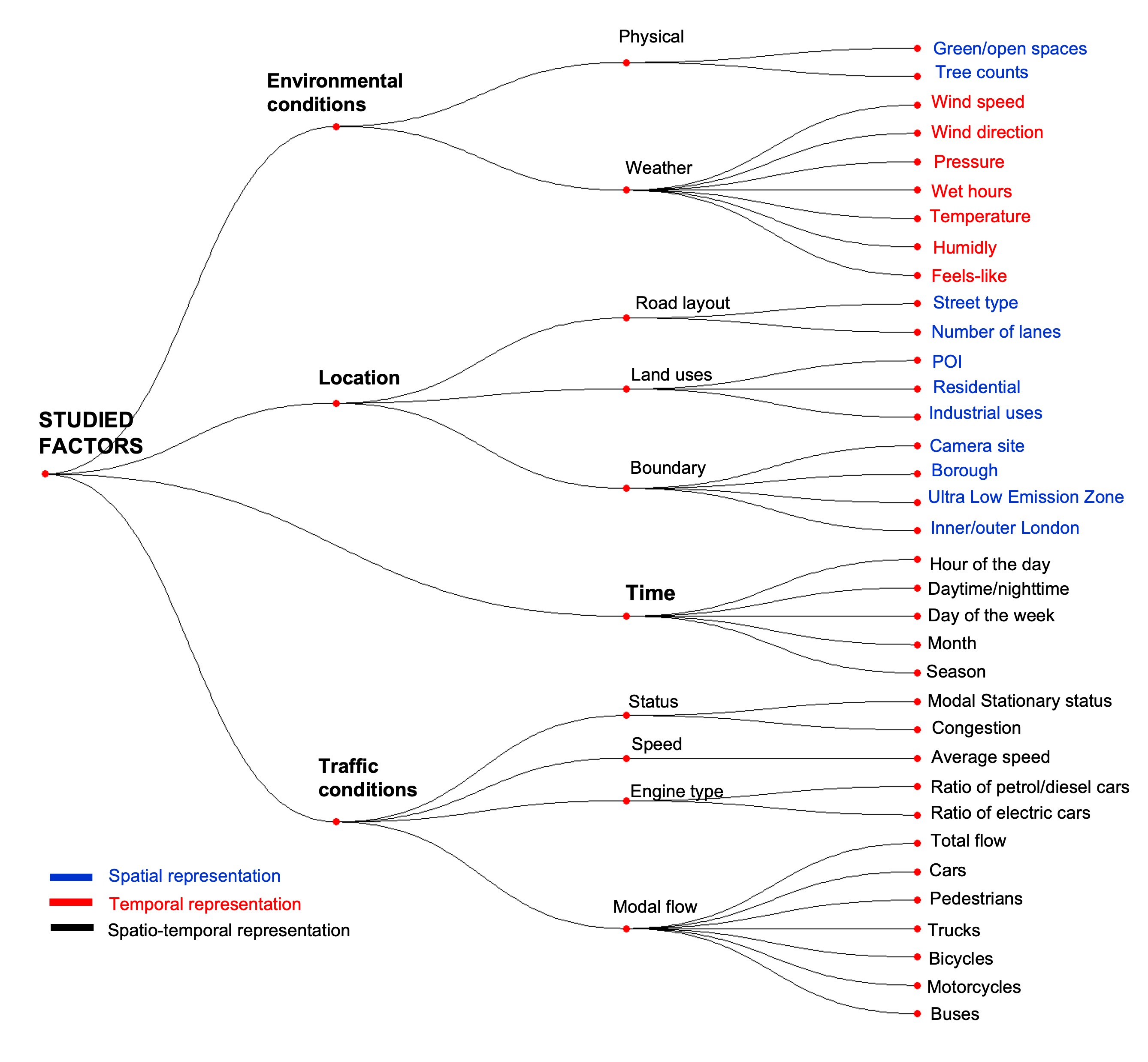}
\caption{Representation of key studied factors and their domain and temporal representation.}
\label{fig:sfig1}
\end{figure*}
\clearpage  

\begin{figure*}[h]
\centering
\includegraphics[width=1\linewidth]{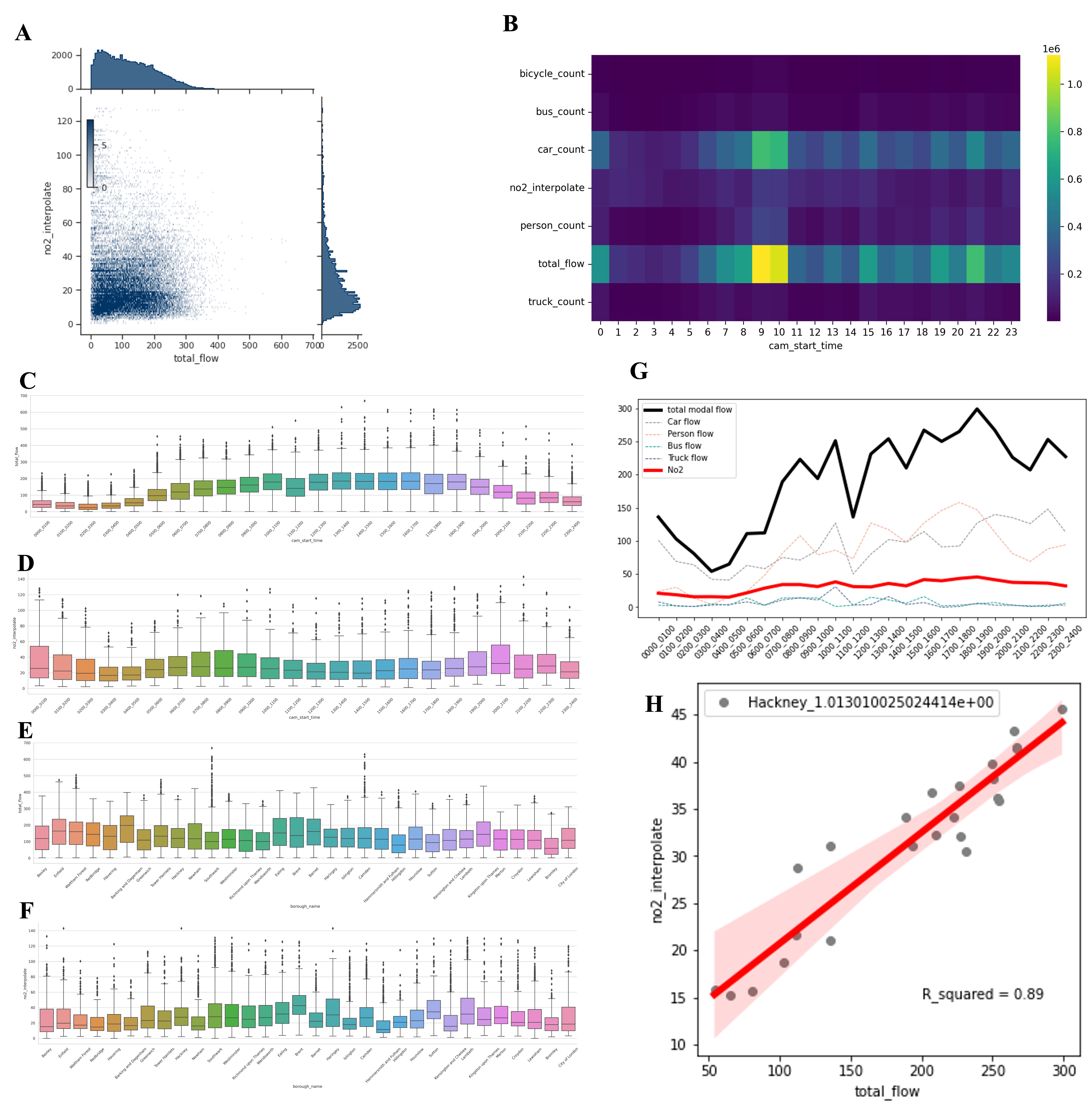}
\caption{Spatiotemporal patterns of traffic flows and NO\textsubscript{2} levels. \textbf{(A)} Relationship between NO\textsubscript{2} levels and total flow for each camera’s location (N=88020). \textbf{(B)} A heatmap showing the relationship between the sum of a given type of traffic modes for all cameras at each hour of the day. \textbf{(C)} A box plot highlighting the distribution of total traffic at a given hour of the day, showing the morning and afternoon peaks. \textbf{(D)} A box plot highlighting the distribution of NO\textsubscript{2} levels at a given hour of the day. \textbf{(E)} A box plot highlighting the distribution of total traffic for each borough in London. \textbf{(F)} A box plot highlighting the distribution of NO\textsubscript{2} levels for each borough in London. \textbf{(G)} The sequence of the different types of traffic modes and NO\textsubscript{2} levels for 24 consecutive hours of a given day and location. \textbf{(H)} The relation between the total traffic modes and NO\textsubscript{2} levels for 24 consecutive hours of a given day and location (n=24, r\textsuperscript{2}=0.89).}
\label{fig:sfig2}
\end{figure*}
\clearpage  

\newpage

\begin{table*}[h]
\centering
\caption{The statistics results of spatial regression model for all hours}
\small  
\begin{tabular}{lrrrr}
\hline
\textbf{Variable} & \textbf{Coeff.} & \textbf{z\_statistics} & \textbf{Std. Error}  & \textbf{P-Value} \\
\hline
CONSTANT & -3240.8 & (-26.593998485571323, 7.964633094485658e-156) & 121.8621 & 0 \\
car\_count & 0.0317 & (15.17594608080966, 5.1036514079524246e-52) & 0.0021 & 0 \\
bus\_count & 0.1212 & (4.968746721566333, 6.738702035838706e-07) & 0.0244 & 0 \\
truck\_count & -0.1463 & (-11.230212099503312, 2.8978208953987843e-29) & 0.013 & 0 \\
motorcycle\_count & -0.6981 & (-1.470587151254405, 0.1414028008301437) & 0.4747 & 0.1414 \\
car\_standing & -0.0847 & (-15.619868885580317, 5.331515286966446e-55) & 0.0054 & 0 \\
bus\_standing & -0.1365 & (-4.490196502306664, 7.115749769009126e-06) & 0.0304 & 0 \\
truck\_standing & 0.0578 & (2.762711349797094, 0.005732343593113537) & 0.0209 & 0.0057 \\
motorcycle\_standing & -1.4161 & (-1.602778838893691, 0.1089834903414405) & 0.8835 & 0.109 \\
congestion & 0.0607 & (16.57835278251283, 9.993053353873868e-62) & 0.0037 & 0 \\
pressure\_mean\_2021 & 3.2438 & (26.962406766252982, 4.080909118979775e-160) & 0.1203 & 0 \\
rainfall\_2021 & 43.571 & (30.49492834300486, 3.0422806498338705e-204) & 1.4288 & 0 \\
sun\_hours\_2021 & 6.4385 & (39.001477963123925, 0.0) & 0.1651 & 0 \\
temperature\_mean\_2021 & -2.4859 & (-53.51203805390859, 0.0) & 0.0465 & 0 \\
wet\_hours\_2021 & -33.7822 & (-35.21633099823809, 1.1246475830658265e-271) & 0.9593 & 0 \\
wind\_speed\_mean\_2021 & 2.8438 & (19.792472757050792, 3.456596745550008e-87) & 0.1437 & 0 \\
Monday & -11.8181 & (-10.519019132738338, 7.060475851963172e-26) & 1.1235 & 0 \\
Tuesday & -76.3571 & (-28.311037979505937, 2.527558312459575e-176) & 2.6971 & 0 \\
Wednesday & -28.0184 & (-17.753650712876322, 1.6150554362803476e-70) & 1.5782 & 0 \\
Thursday & -46.2311 & (-29.82590566064918, 1.8030491978206822e-195) & 1.55 & 0 \\
Friday & -29.808 & (-31.23690461896308, 3.3625209611926207e-214) & 0.9543 & 0 \\
Saturday & -45.0216 & (-33.06182101991166, 1.0520591759555352e-239) & 1.3617 & 0 \\
proximity\_industry & 2.1559 & (11.934337332577172, 7.838059549555e-33) & 0.1806 & 0 \\
wind\_direction\_NE & -26.8775 & (-30.56339577950271, 3.7535466727372635e-205) & 0.8794 & 0 \\
wind\_direction\_SW & 6.761 & (3.911454462153434, 9.17419558745479e-05) & 1.7285 & 0.0001 \\
ultra\_low\_emission\_zone & 3.0747 & (17.030505814120826, 4.878075399459055e-65) & 0.1805 & 0 \\
landuse\_forest & 1.0433 & (8.045252077115027, 8.606807616698688e-16) & 0.1297 & 0 \\
trunk\_street & 1.1702 & (3.5721390778755864, 0.0003540772769421485) & 0.3276 & 0.0004 \\
maxspeed & -0.042 & (-9.303812353819943, 1.354981237316461e-20) & 0.0045 & 0 \\
landuse\_farmland & -0.8048 & (-3.223911552235775, 0.0012645244975738265) & 0.2496 & 0.0013 \\
W\_No2 & 0.0224 & (0.6408633874616294, 0.5216114451415428) & 0.035 & 0.5216 \\
\hline
\end{tabular}
\label{tab:stats_regression}
\end{table*}

\newpage


\begin{longtable}{llrrrr}  
\caption{The statistics results of the spatial regression model for each hour} \\
\hline
\textbf{Hour} & \textbf{Variable} & \textbf{Coeff.} & \textbf{z\_statistics} & \textbf{Std. Error}  & \textbf{P-Value} \\
\hline
\endfirsthead

\hline
\textbf{Hour} & \textbf{Variable} & \textbf{Coeff.} & \textbf{z\_statistics} & \textbf{Std. Error}  & \textbf{P-Value} \\
\hline
\endhead

\hline
\endfoot

\endlastfoot

0000-0100 & CONSTANT & -17.105 & (-1.422813226440614, 0.15479030611761294) & 12.0219 & 0.1548 \\
0000-0100 & car\_count & -0.6192 & (-11.819739627264475, 3.0863961557362466e-32) & 0.0524 & 0 \\
0000-0100 & car\_standing & 0.9992 & (8.532193869768667, 1.4359814784144314e-17) & 0.1171 & 0 \\
0000-0100 & bus\_count & 0.9201 & (1.830824603958303, 0.06712672672985578) & 0.5026 & 0.0671 \\
0000-0100 & bus\_standing & -0.2127 & (-0.3348247173472378, 0.7377573139037313) & 0.6352 & 0.7378 \\
0000-0100 & truck\_count & 1.6024 & (4.626597753966568, 3.717214602491084e-06) & 0.3464 & 0 \\
0000-0100 & truck\_standing & -0.5564 & (-1.2998949896625933, 0.1936369625752341) & 0.428 & 0.1936 \\
0000-0100 & congestion & -0.5106 & (-5.854482869187832, 4.7849652692640265e-09) & 0.0872 & 0 \\
0000-0100 & ultra\_low\_emission\_zone & -1.1388 & (-0.4000536583131751, 0.6891169956944563) & 2.8466 & 0.6891 \\
0000-0100 & proximity\_industry & -1.0366 & (-0.5585707727513226, 0.5764546914629414) & 1.8558 & 0.5765 \\
0000-0100 & maxspeed & 0.0679 & (1.986899300330918, 0.04693355718902318) & 0.0342 & 0.0469 \\
0000-0100 & landuse\_farmland & 1.8319 & (0.6165069992893615, 0.5375599538329356) & 2.9714 & 0.5376 \\
0000-0100 & W\_No2 & 1.7611 & (4.98441349928278, 6.21500415990157e-07) & 0.3533 & 0 \\
0100-0200 & CONSTANT & -50.0638 & (-3.3362236447802345, 0.0008492481546687452) & 15.0061 & 0.0008 \\
0100-0200 & car\_count & -0.7554 & (-11.054180078453335, 2.092399293527026e-28) & 0.0683 & 0 \\
0100-0200 & car\_standing & 1.1265 & (8.051222205107491, 8.197129130942354e-16) & 0.1399 & 0 \\
0100-0200 & bus\_count & 2.176 & (3.77490854403969, 0.0001600663777019936) & 0.5764 & 0.0002 \\
0100-0200 & bus\_standing & -1.2631 & (-1.8105988663391486, 0.07020296782134858) & 0.6976 & 0.0702 \\
0100-0200 & truck\_count & 1.1694 & (3.3709808473980556, 0.0007490106493686409) & 0.3469 & 0.0007 \\
0100-0200 & truck\_standing & -0.3217 & (-0.7274415890747786, 0.466955485158681) & 0.4422 & 0.467 \\
0100-0200 & congestion & -0.4724 & (-4.6549809145614605, 3.2401071888991335e-06) & 0.1015 & 0 \\
0100-0200 & ultra\_low\_emission\_zone & -8.1497 & (-2.5483728599870537, 0.010822672835888104) & 3.198 & 0.0108 \\
0100-0200 & proximity\_industry & 0.9368 & (0.5603078041886621, 0.57526950515043) & 1.6719 & 0.5753 \\
0100-0200 & maxspeed & 0.0962 & (2.607272779901439, 0.009126662120787653) & 0.0369 & 0.0091 \\
0100-0200 & landuse\_farmland & 3.9429 & (1.3218011607964901, 0.18623437171265134) & 2.983 & 0.1862 \\
0100-0200 & W\_No2 & 2.8401 & (5.820974047444298, 5.850565468487407e-09) & 0.4879 & 0 \\
0200-0300 & CONSTANT & 72.6384 & (6.36364438101125, 1.9702201373129983e-10) & 11.4146 & 0 \\
0200-0300 & car\_count & -0.4453 & (-9.725079261257264, 2.357217449034196e-22) & 0.0458 & 0 \\
0200-0300 & car\_standing & 0.5715 & (6.277941271445471, 3.43085449910835e-10) & 0.091 & 0 \\
0200-0300 & bus\_count & -0.8782 & (-2.1305446290287198, 0.03312667537872529) & 0.4122 & 0.0331 \\
0200-0300 & bus\_standing & 1.1046 & (2.182440157801496, 0.029077063275754947) & 0.5061 & 0.0291 \\
0200-0300 & truck\_count & -0.102 & (-0.4883941137369355, 0.6252707104081452) & 0.2089 & 0.6253 \\
0200-0300 & truck\_standing & 0.2373 & (0.8295124361933988, 0.4068145026256976) & 0.2861 & 0.4068 \\
0200-0300 & congestion & -0.3018 & (-4.467035770400603, 7.931087258549044e-06) & 0.0676 & 0 \\
0200-0300 & ultra\_low\_emission\_zone & 10.5295 & (6.8481033828906535, 7.48354785962433e-12) & 1.5376 & 0 \\
0200-0300 & proximity\_industry & 6.2339 & (4.378411929672644, 1.1954723227246431e-05) & 1.4238 & 0 \\
0200-0300 & maxspeed & -0.0408 & (-1.6163579139991175, 0.10601695208724987) & 0.0253 & 0.106 \\
0200-0300 & landuse\_farmland & -7.0769 & (-3.2923528070890624, 0.0009935288968460137) & 2.1495 & 0.001 \\
0200-0300 & W\_No2 & -1.4231 & (-3.276758649009386, 0.0010500608657204522) & 0.4343 & 0.0011 \\
0300-0400 & CONSTANT & 8.5543 & (1.5231444169099424, 0.12772257849910715) & 5.6162 & 0.1277 \\
0300-0400 & car\_count & -0.1192 & (-6.164669383684853, 7.063055580081696e-10) & 0.0193 & 0 \\
0300-0400 & car\_standing & 0.1841 & (3.749933225307289, 0.00017688166523440206) & 0.0491 & 0.0002 \\
0300-0400 & bus\_count & 0.6259 & (3.8484820542389824, 0.0001188520019062718) & 0.1626 & 0.0001 \\
0300-0400 & bus\_standing & -0.2407 & (-1.1170864530977223, 0.26395736216987875) & 0.2155 & 0.264 \\
0300-0400 & truck\_count & 0.6508 & (6.673432185053296, 2.498890829921724e-11) & 0.0975 & 0 \\
0300-0400 & truck\_standing & -0.3339 & (-2.495104256216602, 0.012592012703382227) & 0.1338 & 0.0126 \\
0300-0400 & congestion & -0.0837 & (-2.0799834613307935, 0.03752704985830761) & 0.0402 & 0.0375 \\
0300-0400 & ultra\_low\_emission\_zone & 0.8672 & (1.5029560986843868, 0.1328503652424885) & 0.577 & 0.1329 \\
0300-0400 & proximity\_industry & 1.2278 & (1.4362871895051006, 0.15092063777272793) & 0.8548 & 0.1509 \\
0300-0400 & maxspeed & -0.0148 & (-1.1255001204793111, 0.26037716619101614) & 0.0132 & 0.2604 \\
0300-0400 & landuse\_farmland & -1.1572 & (-1.077205053116908, 0.28138866830289966) & 1.0742 & 0.2814 \\
0300-0400 & W\_No2 & 0.572 & (2.0636688398454637, 0.03904912570403052) & 0.2772 & 0.039 \\
0400-0500 & CONSTANT & 9.6552 & (2.9868384013980154, 0.002818786784857815) & 3.2326 & 0.0028 \\
0400-0500 & car\_count & -0.0989 & (-6.564226012647456, 5.230381902767192e-11) & 0.0151 & 0 \\
0400-0500 & car\_standing & 0.168 & (3.588442829265443, 0.00033265891928589377) & 0.0468 & 0.0003 \\
0400-0500 & bus\_count & 0.3446 & (2.8195201279255087, 0.004809551452660119) & 0.1222 & 0.0048 \\
0400-0500 & bus\_standing & 0.1174 & (0.7427211459173997, 0.45765052462749234) & 0.158 & 0.4577 \\
0400-0500 & truck\_count & 0.6801 & (9.236166425453701, 2.5548697770994693e-20) & 0.0736 & 0 \\
0400-0500 & truck\_standing & -0.3434 & (-3.0118445579026685, 0.0025966552640060187) & 0.114 & 0.0026 \\
0400-0500 & congestion & -0.0543 & (-1.3758593665495125, 0.1688651773696399) & 0.0394 & 0.1689 \\
0400-0500 & ultra\_low\_emission\_zone & 0.7408 & (1.3416845271385838, 0.17969830602222148) & 0.5521 & 0.1797 \\
0400-0500 & proximity\_industry & 2.2431 & (2.869957297428054, 0.004105272361000629) & 0.7816 & 0.0041 \\
0400-0500 & maxspeed & -0.0131 & (-1.0250927600097801, 0.30531942136141643) & 0.0128 & 0.3053 \\
0400-0500 & landuse\_farmland & -0.5522 & (-0.5404555898098896, 0.5888828799520649) & 1.0218 & 0.5889 \\
0400-0500 & W\_No2 & 0.4532 & (2.9392039969924744, 0.0032905641526826722) & 0.1542 & 0.0033 \\
0500-0600 & CONSTANT & 1.4765 & (0.41754705219623617, 0.676278315819081) & 3.5362 & 0.6763 \\
0500-0600 & car\_count & -0.0584 & (-5.356501968193461, 8.484857638299245e-08) & 0.0109 & 0 \\
0500-0600 & car\_standing & 0.0776 & (2.058742188911834, 0.039518942375943215) & 0.0377 & 0.0395 \\
0500-0600 & bus\_count & -0.2824 & (-2.9221258452879093, 0.0034765104067956193) & 0.0966 & 0.0035 \\
0500-0600 & bus\_standing & 0.375 & (2.8766725044951658, 0.00401892459217479) & 0.1303 & 0.004 \\
0500-0600 & truck\_count & 0.1966 & (3.9436237027179883, 8.025956141253998e-05) & 0.0498 & 0.0001 \\
0500-0600 & truck\_standing & -0.0772 & (-0.8835682642354794, 0.3769293169499466) & 0.0873 & 0.3769 \\
0500-0600 & congestion & 0.0217 & (0.6890480340196372, 0.49079304039819716) & 0.0314 & 0.4908 \\
0500-0600 & ultra\_low\_emission\_zone & 0.0455 & (0.07285989935797017, 0.9419176047687764) & 0.6242 & 0.9419 \\
0500-0600 & proximity\_industry & 1.0702 & (1.2479152652200547, 0.21206208971350637) & 0.8576 & 0.2121 \\
0500-0600 & maxspeed & 0.0054 & (0.36430476269294887, 0.7156304437728774) & 0.0147 & 0.7156 \\
0500-0600 & landuse\_farmland & -0.3622 & (-0.3098374170821315, 0.7566845960915674) & 1.1691 & 0.7567 \\
0500-0600 & W\_No2 & 0.9034 & (6.780043086785884, 1.2014004750351273e-11) & 0.1332 & 0 \\
0600-0700 & CONSTANT & 17.9726 & (3.665001893962824, 0.0002473367526108635) & 4.9039 & 0.0002 \\
0600-0700 & car\_count & -0.0284 & (-2.6078568558304926, 0.009111104624793366) & 0.0109 & 0.0091 \\
0600-0700 & car\_standing & -0.0345 & (-1.1223304495109225, 0.261721964116711) & 0.0307 & 0.2617 \\
0600-0700 & bus\_count & 0.0771 & (0.9466551813132724, 0.3438145155086414) & 0.0814 & 0.3438 \\
0600-0700 & bus\_standing & -0.1067 & (-0.9695046602128293, 0.33229345618748163) & 0.1101 & 0.3323 \\
0600-0700 & truck\_count & 0.108 & (2.3194639230645904, 0.020369895101611815) & 0.0466 & 0.0204 \\
0600-0700 & truck\_standing & -0.0845 & (-1.0545185965440356, 0.2916455510697695) & 0.0801 & 0.2916 \\
0600-0700 & congestion & 0.0986 & (4.122920655142562, 3.740985875204968e-05) & 0.0239 & 0 \\
0600-0700 & ultra\_low\_emission\_zone & -0.1285 & (-0.2114741828526926, 0.8325172757614525) & 0.6076 & 0.8325 \\
0600-0700 & proximity\_industry & 2.3754 & (2.7304623344912793, 0.006324555975780188) & 0.87 & 0.0063 \\
0600-0700 & maxspeed & -0.0085 & (-0.5564112327399354, 0.5779297617570607) & 0.0153 & 0.5779 \\
0600-0700 & landuse\_farmland & -1.068 & (-0.8824274455016422, 0.377545699319496) & 1.2102 & 0.3775 \\
0600-0700 & W\_No2 & 0.3035 & (1.8957234376284136, 0.057996624436640776) & 0.1601 & 0.058 \\
0700-0800 & CONSTANT & 12.8064 & (2.3113166533663656, 0.02081537082798931) & 5.5407 & 0.0208 \\
0700-0800 & car\_count & 0.0168 & (1.5312915141914831, 0.12569736185994582) & 0.011 & 0.1257 \\
0700-0800 & car\_standing & -0.1244 & (-4.437282403722675, 9.110173867441231e-06) & 0.028 & 0 \\
0700-0800 & bus\_count & 0.0694 & (0.7660531941133752, 0.44364465155632204) & 0.0906 & 0.4436 \\
0700-0800 & bus\_standing & -0.2261 & (-2.0024537015481725, 0.045235957784405444) & 0.1129 & 0.0452 \\
0700-0800 & truck\_count & 0.0447 & (0.9370720495236022, 0.34872149770087824) & 0.0476 & 0.3487 \\
0700-0800 & truck\_standing & -0.1031 & (-1.3199612889755858, 0.18684794292495743) & 0.0781 & 0.1868 \\
0700-0800 & congestion & 0.1356 & (7.017330756939118, 2.2614682950717765e-12) & 0.0193 & 0 \\
0700-0800 & ultra\_low\_emission\_zone & 1.0992 & (1.5386680830012995, 0.12388534804702023) & 0.7144 & 0.1239 \\
0700-0800 & proximity\_industry & 1.6435 & (1.8670231049527661, 0.06189836137860984) & 0.8803 & 0.0619 \\
0700-0800 & maxspeed & -0.0236 & (-1.3608853281221414, 0.17354993081382286) & 0.0173 & 0.1735 \\
0700-0800 & landuse\_farmland & -0.1782 & (-0.13897973972306754, 0.8894661590191018) & 1.2825 & 0.8895 \\
0700-0800 & W\_No2 & 0.4537 & (2.747752715814393, 0.00600052462852439) & 0.1651 & 0.006 \\
0800-0900 & CONSTANT & -5.4055 & (-0.9387801456674113, 0.34784363464075074) & 5.758 & 0.3478 \\
0800-0900 & car\_count & -0.0616 & (-4.052649560525262, 5.064082967360317e-05) & 0.0152 & 0.0001 \\
0800-0900 & car\_standing & 0.1361 & (3.3412709332109403, 0.0008339580002075611) & 0.0407 & 0.0008 \\
0800-0900 & bus\_count & -0.1098 & (-0.8587524741950237, 0.39047709310566614) & 0.1279 & 0.3905 \\
0800-0900 & bus\_standing & 0.3701 & (2.256877772684688, 0.02401571329393297) & 0.164 & 0.024 \\
0800-0900 & truck\_count & 0.1853 & (2.7706509455543977, 0.00559443632625586) & 0.0669 & 0.0056 \\
0800-0900 & truck\_standing & 0.3434 & (2.936680241670709, 0.0033174597338085344) & 0.1169 & 0.0033 \\
0800-0900 & congestion & -0.1368 & (-4.476212845534083, 7.597874091528457e-06) & 0.0306 & 0 \\
0800-0900 & ultra\_low\_emission\_zone & -1.4664 & (-1.3070998709646098, 0.19117879511132674) & 1.1219 & 0.1912 \\
0800-0900 & proximity\_industry & 0.5013 & (0.4411851257607863, 0.6590789805228792) & 1.1362 & 0.6591 \\
0800-0900 & maxspeed & 0.0082 & (0.3379695107089135, 0.735386164170585) & 0.0244 & 0.7354 \\
0800-0900 & landuse\_farmland & -0.2146 & (-0.11609442884086187, 0.9075777041003068) & 1.8489 & 0.9076 \\
0800-0900 & W\_No2 & 1.2653 & (7.263437860150474, 3.7737384219405366e-13) & 0.1742 & 0 \\
0900-1000 & CONSTANT & -6.6706 & (-1.3666684363326176, 0.17172925557214314) & 4.8809 & 0.1717 \\
0900-1000 & car\_count & -0.0614 & (-5.048757714021111, 4.4469228379809145e-07) & 0.0122 & 0 \\
0900-1000 & car\_standing & 0.1293 & (4.196557903740315, 2.7100219941570485e-05) & 0.0308 & 0 \\
0900-1000 & bus\_count & -0.2764 & (-2.5596030434505606, 0.010479178430431602) & 0.108 & 0.0105 \\
0900-1000 & bus\_standing & 0.5356 & (3.91673813898755, 8.975515074887329e-05) & 0.1367 & 0.0001 \\
0900-1000 & truck\_count & 0.2637 & (4.744172839407008, 2.0936003858691974e-06) & 0.0556 & 0 \\
0900-1000 & truck\_standing & 0.0733 & (0.8289021884558828, 0.4071597569305112) & 0.0884 & 0.4072 \\
0900-1000 & congestion & -0.1237 & (-5.728264537278641, 1.0146325948632018e-08) & 0.0216 & 0 \\
0900-1000 & ultra\_low\_emission\_zone & -2.4566 & (-2.1521341709414576, 0.031386788515926155) & 1.1415 & 0.0314 \\
0900-1000 & proximity\_industry & 0.6393 & (0.6622332392391975, 0.5078217549131003) & 0.9653 & 0.5078 \\
0900-1000 & maxspeed & 0.0196 & (0.9493548421218491, 0.34244016947766664) & 0.0207 & 0.3424 \\
0900-1000 & landuse\_farmland & -1.6415 & (-1.0588016498164563, 0.2896901221510638) & 1.5504 & 0.2897 \\
0900-1000 & W\_No2 & 1.3439 & (8.70569834213755, 3.156252849345412e-18) & 0.1544 & 0 \\
1000-1100 & CONSTANT & -0.3092 & (-0.08230152486511495, 0.9344069418690983) & 3.7564 & 0.9344 \\
1000-1100 & car\_count & -0.0595 & (-5.103712513647672, 3.3305411184784755e-07) & 0.0117 & 0 \\
1000-1100 & car\_standing & 0.126 & (4.194886253925344, 2.7300855537905636e-05) & 0.03 & 0 \\
1000-1100 & bus\_count & 0.1508 & (1.3485831927421588, 0.17747088214903417) & 0.1118 & 0.1775 \\
1000-1100 & bus\_standing & 0.0861 & (0.6137524733784656, 0.5393789076303346) & 0.1403 & 0.5394 \\
1000-1100 & truck\_count & 0.2977 & (5.11109651069052, 3.20294295750069e-07) & 0.0582 & 0 \\
1000-1100 & truck\_standing & 0.0248 & (0.27266522360955026, 0.7851105695931626) & 0.0909 & 0.7851 \\
1000-1100 & congestion & -0.1173 & (-5.644581736288564, 1.6558330218078854e-08) & 0.0208 & 0 \\
1000-1100 & ultra\_low\_emission\_zone & -1.1356 & (-1.241948144622478, 0.21425569510412268) & 0.9144 & 0.2143 \\
1000-1100 & proximity\_industry & 0.4662 & (0.4834725202351828, 0.6287602668798796) & 0.9643 & 0.6288 \\
1000-1100 & maxspeed & 0.0079 & (0.3891923478974078, 0.6971338636137767) & 0.0203 & 0.6971 \\
1000-1100 & landuse\_farmland & 0.1096 & (0.06942963962586732, 0.9446476368279135) & 1.5787 & 0.9446 \\
1000-1100 & W\_No2 & 1.1147 & (8.790363555211073, 1.4907676312450971e-18) & 0.1268 & 0 \\
1100-1200 & CONSTANT & 5.7964 & (1.5810402678554063, 0.11386883086647294) & 3.6662 & 0.1139 \\
1100-1200 & car\_count & -0.0274 & (-2.5832917699977154, 0.00978625221458549) & 0.0106 & 0.0098 \\
1100-1200 & car\_standing & 0.0291 & (1.0479700029947427, 0.2946524276155944) & 0.0278 & 0.2947 \\
1100-1200 & bus\_count & 0.3441 & (2.8713224337877876, 0.004087583551811111) & 0.1198 & 0.0041 \\
1100-1200 & bus\_standing & -0.1679 & (-1.1419501638409777, 0.253474735465701) & 0.147 & 0.2535 \\
1100-1200 & truck\_count & 0.4358 & (6.9098519091475294, 4.851599044974655e-12) & 0.0631 & 0 \\
1100-1200 & truck\_standing & -0.1215 & (-1.2285229696970534, 0.21925071215303749) & 0.0989 & 0.2193 \\
1100-1200 & congestion & -0.0108 & (-0.5931620021394591, 0.55307273951077) & 0.0183 & 0.5531 \\
1100-1200 & ultra\_low\_emission\_zone & 2.1103 & (1.972578675196157, 0.04854358455823267) & 1.0698 & 0.0485 \\
1100-1200 & proximity\_industry & 0.3068 & (0.37356692833626337, 0.7087265427157636) & 0.8213 & 0.7087 \\
1100-1200 & maxspeed & -0.0302 & (-1.7383580118643138, 0.08214775030689517) & 0.0174 & 0.0821 \\
1100-1200 & landuse\_farmland & -0.2346 & (-0.1721050691235831, 0.8633549276451437) & 1.3634 & 0.8634 \\
1100-1200 & W\_No2 & 0.6331 & (4.770358795146408, 1.8389806139686912e-06) & 0.1327 & 0 \\
1200-1300 & CONSTANT & -3.3924 & (-1.4634658078082978, 0.14333996305211502) & 2.318 & 0.1433 \\
1200-1300 & car\_count & -0.0191 & (-2.3953742651391434, 0.016603407735358212) & 0.008 & 0.0166 \\
1200-1300 & car\_standing & 0.0386 & (1.9936763482326816, 0.046187435748492114) & 0.0194 & 0.0462 \\
1200-1300 & bus\_count & -0.0265 & (-0.2957731805888858, 0.7674033058827321) & 0.0896 & 0.7674 \\
1200-1300 & bus\_standing & 0.114 & (1.0496951228940687, 0.29385830679146263) & 0.1086 & 0.2939 \\
1200-1300 & truck\_count & 0.0895 & (2.0479791439056343, 0.04056204302010602) & 0.0437 & 0.0406 \\
1200-1300 & truck\_standing & 0.1167 & (1.7086854858126603, 0.08750922406530509) & 0.0683 & 0.0875 \\
1200-1300 & motorcycle\_count & 0.4211 & (1.1387970654382311, 0.25478780607461293) & 0.3697 & 0.2548 \\
1200-1300 & motorcycle\_standing & -1.2002 & (-1.7183699137118276, 0.08572916446165056) & 0.6985 & 0.0857 \\
1200-1300 & congestion & -0.0379 & (-2.961420386890992, 0.003062236505885295) & 0.0128 & 0.0031 \\
1200-1300 & ultra\_low\_emission\_zone & -0.8467 & (-1.2273614474044903, 0.2196867697715057) & 0.6899 & 0.2197 \\
1200-1300 & proximity\_industry & 0.4641 & (0.7071318232890271, 0.4794845613345169) & 0.6563 & 0.4795 \\
1200-1300 & maxspeed & 0.0037 & (0.27200255217383623, 0.7856200569576068) & 0.0135 & 0.7856 \\
1200-1300 & landuse\_farmland & 0.5362 & (0.4891707854437782, 0.6247207907265608) & 1.096 & 0.6247 \\
1200-1300 & W\_No2 & 1.1575 & (12.666143067586617, 9.10838592442768e-37) & 0.0914 & 0 \\
1300-1400 & CONSTANT & -0.2198 & (-0.06839979108685945, 0.9454673881309539) & 3.2135 & 0.9455 \\
1300-1400 & car\_count & -0.0276 & (-2.8677368337176707, 0.0041341925438972225) & 0.0096 & 0.0041 \\
1300-1400 & car\_standing & 0.0285 & (1.2398933841673851, 0.21501483124028797) & 0.023 & 0.215 \\
1300-1400 & bus\_count & 0.0111 & (0.10842127313806564, 0.9136615270790855) & 0.1024 & 0.9137 \\
1300-1400 & bus\_standing & 0.1285 & (1.032202284271459, 0.3019773651899743) & 0.1245 & 0.302 \\
1300-1400 & truck\_count & 0.2187 & (4.213756179342229, 2.5115841426167324e-05) & 0.0519 & 0 \\
1300-1400 & truck\_standing & 0.0325 & (0.3900404077645584, 0.6965066674250593) & 0.0832 & 0.6965 \\
1300-1400 & congestion & -0.0308 & (-2.0375636974736526, 0.041593585658042) & 0.0151 & 0.0416 \\
1300-1400 & ultra\_low\_emission\_zone & -0.1218 & (-0.14987906003154955, 0.8808600327604327) & 0.8124 & 0.8809 \\
1300-1400 & proximity\_industry & 0.3506 & (0.42447119551748286, 0.6712222079125278) & 0.826 & 0.6712 \\
1300-1400 & maxspeed & -0.001 & (-0.0593849739131664, 0.9526454810168881) & 0.0173 & 0.9526 \\
1300-1400 & landuse\_farmland & 0.539 & (0.39353531464565283, 0.6939241463458214) & 1.3697 & 0.6939 \\
1300-1400 & W\_No2 & 1.0066 & (8.312699196446937, 9.354943568784711e-17) & 0.1211 & 0 \\
1400-1500 & CONSTANT & -0.7632 & (-0.18285576579534385, 0.8549111914045626) & 4.1739 & 0.8549 \\
1400-1500 & car\_count & -0.0134 & (-1.7217208522370624, 0.0851200980900606) & 0.0078 & 0.0851 \\
1400-1500 & car\_standing & 0.0136 & (0.7628140410396775, 0.44557430608543647) & 0.0179 & 0.4456 \\
1400-1500 & bus\_count & 0.2705 & (3.138606593407384, 0.001697531821176227) & 0.0862 & 0.0017 \\
1400-1500 & bus\_standing & -0.1871 & (-1.7989653574024034, 0.07202416081184737) & 0.104 & 0.072 \\
1400-1500 & truck\_count & 0.17 & (3.739538472242519, 0.00018435843189517522) & 0.0455 & 0.0002 \\
1400-1500 & truck\_standing & 0.053 & (0.7220524740185159, 0.47026221630366305) & 0.0734 & 0.4703 \\
1400-1500 & congestion & -0.0185 & (-1.6100048121506372, 0.10739680576195645) & 0.0115 & 0.1074 \\
1400-1500 & ultra\_low\_emission\_zone & -0.1167 & (-0.12873370124897857, 0.8975683675386823) & 0.9064 & 0.8976 \\
1400-1500 & proximity\_industry & 0.2375 & (0.3413623673705277, 0.7328308019381293) & 0.6957 & 0.7328 \\
1400-1500 & maxspeed & -0.0155 & (-1.1514698695521437, 0.24953898396970786) & 0.0135 & 0.2495 \\
1400-1500 & landuse\_farmland & 0.448 & (0.37976738971286506, 0.7041180912024858) & 1.1798 & 0.7041 \\
1400-1500 & W\_No2 & 0.9862 & (5.943523942893663, 2.7895898423099583e-09) & 0.1659 & 0 \\
1500-1600 & CONSTANT & 0.089 & (0.02851345030969469, 0.9772526405955085) & 3.1201 & 0.9773 \\
1500-1600 & car\_count & -0.0068 & (-0.6997786286498803, 0.48406556330665207) & 0.0098 & 0.4841 \\
1500-1600 & car\_standing & -0.0299 & (-1.2416377421873876, 0.21437024991724507) & 0.0241 & 0.2144 \\
1500-1600 & bus\_count & 0.3122 & (2.5181268450442498, 0.011798082386365505) & 0.124 & 0.0118 \\
1500-1600 & bus\_standing & -0.2818 & (-1.8918961829846315, 0.058504814464961585) & 0.149 & 0.0585 \\
1500-1600 & truck\_count & 0.2355 & (3.559799593303441, 0.000371137913473967) & 0.0661 & 0.0004 \\
1500-1600 & truck\_standing & 0.0807 & (0.7911954275581424, 0.428829959316263) & 0.1019 & 0.4288 \\
1500-1600 & congestion & 0.0038 & (0.23615447947094417, 0.8133128006772714) & 0.0162 & 0.8133 \\
1500-1600 & ultra\_low\_emission\_zone & -0.075 & (-0.09126109421142045, 0.9272851312625744) & 0.8219 & 0.9273 \\
1500-1600 & proximity\_industry & 0.2292 & (0.2506122933029184, 0.8021138763059137) & 0.9146 & 0.8021 \\
1500-1600 & maxspeed & -0.0129 & (-0.6924173553426339, 0.4886752698470491) & 0.0187 & 0.4887 \\
1500-1600 & landuse\_farmland & 0.1686 & (0.11474484011885955, 0.9086473713538677) & 1.4692 & 0.9086 \\
1500-1600 & W\_No2 & 0.9329 & (8.604338188369379, 7.675835279392488e-18) & 0.1084 & 0 \\
1600-1700 & CONSTANT & -0.8019 & (-0.174394225637206, 0.8615556520613307) & 4.598 & 0.8616 \\
1600-1700 & car\_count & 0.0132 & (1.4385980396276223, 0.15026444211465523) & 0.0092 & 0.1503 \\
1600-1700 & car\_standing & -0.0539 & (-2.5192382519531753, 0.011760904653262108) & 0.0214 & 0.0118 \\
1600-1700 & bus\_count & 0.0354 & (0.30629868652826564, 0.7593772391883085) & 0.1157 & 0.7594 \\
1600-1700 & bus\_standing & 0.0843 & (0.5995016685302481, 0.54883839775423) & 0.1406 & 0.5488 \\
1600-1700 & truck\_count & 0.3613 & (4.679428788996316, 2.8767525130330154e-06) & 0.0772 & 0 \\
1600-1700 & truck\_standing & -0.0822 & (-0.725978419177339, 0.46785200050772824) & 0.1133 & 0.4679 \\
1600-1700 & congestion & 0.0344 & (2.4495655104317264, 0.014302868895399464) & 0.014 & 0.0143 \\
1600-1700 & ultra\_low\_emission\_zone & 0.4722 & (0.4583128394197986, 0.6467277010798711) & 1.0303 & 0.6467 \\
1600-1700 & proximity\_industry & 0.4096 & (0.4347159505657996, 0.6637686191140797) & 0.9423 & 0.6638 \\
1600-1700 & maxspeed & -0.0211 & (-1.1682485342554503, 0.24270652841106344) & 0.018 & 0.2427 \\
1600-1700 & landuse\_farmland & 0.202 & (0.14748258473519207, 0.8827511247816592) & 1.3695 & 0.8828 \\
1600-1700 & W\_No2 & 0.8451 & (5.468728639817003, 4.532750840006737e-08) & 0.1545 & 0 \\
1700-1800 & CONSTANT & 53.1597 & (7.592661935949261, 3.1339943993941106e-14) & 7.0015 & 0 \\
1700-1800 & car\_count & -0.0409 & (-4.675501178143865, 2.932365936663897e-06) & 0.0087 & 0 \\
1700-1800 & car\_standing & -0.0669 & (-3.007819725124115, 0.0026312917616127984) & 0.0223 & 0.0026 \\
1700-1800 & bus\_count & 0.3323 & (2.7896030295562655, 0.0052772702631175865) & 0.1191 & 0.0053 \\
1700-1800 & bus\_standing & -0.2328 & (-1.6234636454457498, 0.10449032878954957) & 0.1434 & 0.1045 \\
1700-1800 & truck\_count & 0.3243 & (3.6689159845727097, 0.00024358110416401865) & 0.0884 & 0.0002 \\
1700-1800 & truck\_standing & -0.154 & (-1.2377837751979737, 0.21579624603809455) & 0.1244 & 0.2158 \\
1700-1800 & congestion & 0.0444 & (3.090512769921472, 0.001998112124808064) & 0.0144 & 0.002 \\
1700-1800 & ultra\_low\_emission\_zone & -0.3135 & (-0.47998158728139223, 0.6312404857311082) & 0.6532 & 0.6312 \\
1700-1800 & proximity\_industry & 4.1624 & (4.977733028358603, 6.433330655400384e-07) & 0.8362 & 0 \\
1700-1800 & maxspeed & -0.0275 & (-1.6488622849540624, 0.09917585125303613) & 0.0167 & 0.0992 \\
1700-1800 & landuse\_farmland & 1.1012 & (0.8525393597861025, 0.3939148030444227) & 1.2917 & 0.3939 \\
1700-1800 & W\_No2 & -0.8412 & (-3.474864394437075, 0.0005111116551243819) & 0.2421 & 0.0005 \\
1800-1900 & CONSTANT & 9.3114 & (1.7254280002318938, 0.08445036885771898) & 5.3966 & 0.0845 \\
1800-1900 & car\_count & -0.0169 & (-2.3915749637536403, 0.016776257639071146) & 0.0071 & 0.0168 \\
1800-1900 & car\_standing & -0.0024 & (-0.13328354219144592, 0.8939691428976082) & 0.0182 & 0.894 \\
1800-1900 & bus\_count & 0.3839 & (4.1705471902817965, 3.0386912020756876e-05) & 0.092 & 0 \\
1800-1900 & bus\_standing & -0.2703 & (-2.3856995970016928, 0.01704666921397422) & 0.1133 & 0.017 \\
1800-1900 & truck\_count & 0.4356 & (5.812841075481985, 6.142136777492808e-09) & 0.0749 & 0 \\
1800-1900 & truck\_standing & -0.2157 & (-2.062086568579402, 0.03919949487039771) & 0.1046 & 0.0392 \\
1800-1900 & congestion & 0.0091 & (0.7464509267862569, 0.4553950651988873) & 0.0122 & 0.4554 \\
1800-1900 & ultra\_low\_emission\_zone & 0.3204 & (0.5404066790357264, 0.5889166027277186) & 0.5929 & 0.5889 \\
1800-1900 & proximity\_industry & 1.3496 & (1.7278547297278983, 0.08401427230270565) & 0.7811 & 0.084 \\
1800-1900 & maxspeed & -0.0224 & (-1.646444335189854, 0.09967231159504074) & 0.0136 & 0.0997 \\
1800-1900 & landuse\_farmland & -0.3608 & (-0.328303535891946, 0.7426821709572362) & 1.099 & 0.7427 \\
1800-1900 & W\_No2 & 0.6129 & (3.579921154115987, 0.0003436978969211541) & 0.1712 & 0.0003 \\
1900-2000 & CONSTANT & 31.0755 & (4.912530276976067, 8.990849395047374e-07) & 6.3258 & 0 \\
1900-2000 & car\_count & -0.0184 & (-1.4383910630621797, 0.15032312692574973) & 0.0128 & 0.1503 \\
1900-2000 & car\_standing & -0.0218 & (-0.6277132353310695, 0.5301918134361415) & 0.0347 & 0.5302 \\
1900-2000 & bus\_count & 0.8547 & (4.953921179485579, 7.273267810405693e-07) & 0.1725 & 0 \\
1900-2000 & bus\_standing & -0.5504 & (-2.4709316923382203, 0.013476155702296546) & 0.2228 & 0.0135 \\
1900-2000 & truck\_count & 0.9442 & (7.070584592028268, 1.5428234999852982e-12) & 0.1335 & 0 \\
1900-2000 & truck\_standing & -0.5063 & (-2.6314721513160437, 0.008501583439042705) & 0.1924 & 0.0085 \\
1900-2000 & congestion & 0.027 & (1.1594784795930473, 0.24626120359949777) & 0.0233 & 0.2463 \\
1900-2000 & ultra\_low\_emission\_zone & 4.2567 & (3.5236674641041845, 0.00042561804299153904) & 1.208 & 0.0004 \\
1900-2000 & proximity\_industry & 4.9501 & (3.602291798311308, 0.0003154240094808577) & 1.3741 & 0.0003 \\
1900-2000 & maxspeed & -0.039 & (-1.7435724787579818, 0.08123366573896032) & 0.0224 & 0.0812 \\
1900-2000 & landuse\_farmland & -3.1009 & (-1.6587389951111593, 0.09716839796606679) & 1.8694 & 0.0972 \\
1900-2000 & W\_No2 & -0.0921 & (-0.4972835553194891, 0.6189891066113127) & 0.1851 & 0.619 \\
2000-2100 & CONSTANT & 21.7704 & (3.2628926433028522, 0.0011028128769828723) & 6.6721 & 0.0011 \\
2000-2100 & car\_count & -0.0011 & (-0.07906857235607473, 0.936978080853577) & 0.0142 & 0.937 \\
2000-2100 & car\_standing & -0.002 & (-0.056047484277131146, 0.955303979601727) & 0.0365 & 0.9553 \\
2000-2100 & bus\_count & 0.7195 & (3.4170788225345476, 0.0006329695035492161) & 0.2106 & 0.0006 \\
2000-2100 & bus\_standing & -0.4378 & (-1.636487180728775, 0.10173766646972411) & 0.2675 & 0.1017 \\
2000-2100 & truck\_count & 0.8405 & (5.397351721254542, 6.76317260873408e-08) & 0.1557 & 0 \\
2000-2100 & truck\_standing & -0.6107 & (-2.8609366724401823, 0.004223914150747974) & 0.2135 & 0.0042 \\
2000-2100 & congestion & 0.0008 & (0.032546273682646155, 0.9740364144954817) & 0.0247 & 0.974 \\
2000-2100 & ultra\_low\_emission\_zone & 4.7215 & (3.138990741284355, 0.0016953080427577288) & 1.5041 & 0.0017 \\
2000-2100 & proximity\_industry & 3.3943 & (2.548665180634531, 0.010813606344284098) & 1.3318 & 0.0108 \\
2000-2100 & maxspeed & -0.0341 & (-1.4998917835909071, 0.13364243667114942) & 0.0227 & 0.1336 \\
2000-2100 & landuse\_farmland & -0.6815 & (-0.37585785332604416, 0.7070225750833969) & 1.8131 & 0.707 \\
2000-2100 & W\_No2 & 0.2953 & (1.5893571933067356, 0.1119797731362686) & 0.1858 & 0.112 \\
2100-2200 & CONSTANT & 38.8243 & (6.843206722810891, 7.743977410731256e-12) & 5.6734 & 0 \\
2100-2200 & car\_count & 0.1414 & (7.919058838836297, 2.3931508918998777e-15) & 0.0179 & 0 \\
2100-2200 & car\_standing & -0.228 & (-4.940551198962256, 7.790203314362169e-07) & 0.0462 & 0 \\
2100-2200 & bus\_count & 0.3728 & (1.2786527564327461, 0.20101936318231073) & 0.2916 & 0.201 \\
2100-2200 & bus\_standing & -0.2097 & (-0.5711179211033951, 0.5679197113236731) & 0.3672 & 0.5679 \\
2100-2200 & truck\_count & 0.6375 & (2.831888246391382, 0.00462740183264279) & 0.2251 & 0.0046 \\
2100-2200 & truck\_standing & -0.3301 & (-1.1031170733211604, 0.2699763273615098) & 0.2992 & 0.27 \\
2100-2200 & congestion & 0.1505 & (4.801880225194737, 1.5718270622080728e-06) & 0.0313 & 0 \\
2100-2200 & ultra\_low\_emission\_zone & 7.1217 & (5.261604416706569, 1.428037519751192e-07) & 1.3535 & 0 \\
2100-2200 & proximity\_industry & 9.4025 & (6.045973744272375, 1.4851028193316824e-09) & 1.5552 & 0 \\
2100-2200 & maxspeed & -0.0183 & (-0.8082849163457263, 0.4189265836648054) & 0.0226 & 0.4189 \\
2100-2200 & landuse\_farmland & -1.861 & (-0.989436267333816, 0.3224497375588351) & 1.8809 & 0.3224 \\
2100-2200 & W\_No2 & -0.7097 & (-3.563292453116102, 0.00036623227300334766) & 0.1992 & 0.0004 \\
2200-2300 & CONSTANT & -3.3004 & (-0.8042824617354022, 0.42123386413575203) & 4.1035 & 0.4212 \\
2200-2300 & car\_count & -0.0601 & (-4.618812497820525, 3.859424058457775e-06) & 0.013 & 0 \\
2200-2300 & car\_standing & 0.0623 & (1.894252102930146, 0.05819155526750937) & 0.0329 & 0.0582 \\
2200-2300 & bus\_count & 0.0248 & (0.13307197906809493, 0.8941364554939463) & 0.1862 & 0.8941 \\
2200-2300 & bus\_standing & 0.1058 & (0.4668347550738804, 0.6406181080269093) & 0.2267 & 0.6406 \\
2200-2300 & truck\_count & 0.4878 & (3.4412887497968527, 0.0005789503482378232) & 0.1417 & 0.0006 \\
2200-2300 & truck\_standing & -0.3641 & (-1.8846008395213232, 0.0594837488895432) & 0.1932 & 0.0595 \\
2200-2300 & congestion & -0.0444 & (-1.9893272148625123, 0.046665097663287865) & 0.0223 & 0.0467 \\
2200-2300 & ultra\_low\_emission\_zone & -0.1148 & (-0.15518557036540953, 0.8766750236561937) & 0.7396 & 0.8767 \\
2200-2300 & proximity\_industry & -0.0947 & (-0.10511146144970214, 0.916287364764925) & 0.9009 & 0.9163 \\
2200-2300 & maxspeed & 0.0068 & (0.4389055671089385, 0.660729962700175) & 0.0156 & 0.6607 \\
2200-2300 & landuse\_farmland & 0.4318 & (0.33018998541754424, 0.7412564133809527) & 1.3078 & 0.7413 \\
2200-2300 & W\_No2 & 1.175 & (8.833442354179356, 1.0150265451114971e-18) & 0.133 & 0 \\
2300-2400 & CONSTANT & 2.0293 & (0.5826549420596769, 0.5601256112396158) & 3.4828 & 0.5601 \\
2300-2400 & car\_count & -0.0215 & (-1.6503829076147054, 0.09886464476727462) & 0.013 & 0.0989 \\
2300-2400 & car\_standing & -0.01 & (-0.2813644550450018, 0.7784308780752109) & 0.0354 & 0.7784 \\
2300-2400 & bus\_count & -0.0029 & (-0.01753820282184784, 0.98600725608255) & 0.165 & 0.986 \\
2300-2400 & bus\_standing & -0.0381 & (-0.1774146984156262, 0.8591826627874374) & 0.2147 & 0.8592 \\
2300-2400 & truck\_count & 0.1764 & (1.3614764108283581, 0.17336318322166921) & 0.1296 & 0.1734 \\
2300-2400 & truck\_standing & -0.032 & (-0.1801585450564471, 0.8570281021224445) & 0.1777 & 0.857 \\
2300-2400 & congestion & 0.0181 & (0.7021424233338999, 0.48259034800831413) & 0.0258 & 0.4826 \\
2300-2400 & ultra\_low\_emission\_zone & 0.4688 & (0.7248228923302461, 0.4685606898228226) & 0.6468 & 0.4686 \\
2300-2400 & proximity\_industry & 0.6385 & (0.8773662325801423, 0.3802877480286384) & 0.7278 & 0.3803 \\
2300-2400 & maxspeed & 0.0065 & (0.48301920222555284, 0.6290821019619306) & 0.0134 & 0.6291 \\
2300-2400 & landuse\_farmland & -0.0494 & (-0.04354357043845299, 0.9652682332560485) & 1.1344 & 0.9653 \\
2300-2400 & W\_No2 & 0.9075 & (6.060605791634237, 1.3560982202380852e-09) & 0.1497 & 0 \\
\hline
\end{longtable}

\begin{sidewaystable}
\centering
\caption{Results of Granger causality analysis for all factors in relation to NO\textsubscript{2} levels. 
Symbols for tests: \( \alpha \): SSR F-test, \( \beta \): SSR Chi\textsuperscript{2} test, \( \gamma \): Likelihood ratio test, \( \delta \): Parameter F-test}
\scriptsize  
\begin{tabular}{llrrrrrr}
\hline
\textbf{Variable} & \textbf{CAM ID} & \textbf{Tests} & \textbf{Lag 1} & \textbf{Lag 2} & \textbf{Lag 3} & \textbf{Lag 4} & \textbf{Lag 5} \\
\hline
Car count & Hackney – 1.01301002 & \( \alpha \) & 0.0521 (p0.8219) & 2.2857 (p0.1339) & 1.5208 (p0.2559) & 3.5639 (p0.0469) & 2.5550 (p0.1267) \\
 & (index=2) & \( \beta \) & 0.0603 (p0.8060) & 5.9998 (p0.0498) & 7.0189 (p0.0713) & 27.0857 (p0.0000) & 32.8494 (p0.0000) \\
 &  & \( \gamma \) & 0.0603 (p0.8061) & 5.2775 (p0.0715) & 6.0161 (p0.1108) & 16.8352 (p0.0021) & 18.6929 (p0.0022) \\
 &  & \( \delta \) & 0.0521 (p0.8219) & 2.2857 (p0.1339) & 1.5208 (p0.2559) & 3.5639 (p0.0469) & 2.5550 (p0.1267) \\
Truck (standing) & Hackney – 1.01301002 & \( \alpha \) & 6.0843 (p0.0233) & 2.5576 (p0.1087) & 1.3949 (p0.2886) & 1.0889 (p0.4128) & 1.2504 (p0.3794) \\
 & (index=2) & \( \beta \) & 7.0450 (p0.0079) & 6.7137 (p0.0348) & 6.4382 (p0.0921) & 8.2756 (p0.0820) & 16.0772 (p0.0066) \\
 &  & \( \gamma \) & 6.1117 (p0.0134) & 5.8255 (p0.0543) & 5.5816 (p0.1338) & 6.8695 (p0.1429) & 11.4886 (p0.0425) \\
 &  & \( \delta \) & 6.0843 (p0.0233) & 2.5576 (p0.1087) & 1.3949 (p0.2886) & 1.0889 (p0.4128) & 1.2504 (p0.3794) \\
Truck (standing) & Hackney-1.01406002 & \( \alpha \) & 6.1217 (p0.0230) & 5.4456 (p0.0157) & 3.6537 (p0.0415) & 1.5187 (p0.2692) & 1.0317 (p0.4669) \\
 & (index=8) & \( \beta \) & 7.0883 (p0.0078) & 14.2946 (p0.0008) & 16.8632 (p0.0008) & 11.5418 (p0.0211) & 13.2642 (p0.0210) \\
 &  & \( \gamma \) & 6.1445 (p0.0132) & 10.9034 (p0.0043) & 12.2296 (p0.0066) & 9.0185 (p0.0606) & 9.9378 (p0.0770) \\
 &  & \( \delta \) & 6.1217 (p0.0230) & 5.4456 (p0.0157) & 3.6537 (p0.0415) & 1.5187 (p0.2692) & 1.0317 (p0.4669) \\
Truck count & Hackney – 1.01403999 & \( \alpha \) & 6.4913 (p0.0197) & 3.9238 (p0.0411) & 3.4478 (p0.0486) & 4.0535 (p0.0331) & 3.4907 (p0.0669) \\
 & (index=7) & \( \beta \) & 7.5162 (p0.0061) & 10.3000 (p0.0058) & 15.9129 (p0.0012) & 30.8069 (p0.0000) & 44.8799 (p0.0000) \\
 &  & \( \gamma \) & 6.4657 (p0.0110) & 8.3810 (p0.0151) & 11.7073 (p0.0085) & 18.3106 (p0.0011) & 22.5154 (p0.0004) \\
 &  & \( \delta \) & 6.4913 (p0.0197) & 3.9238 (p0.0411) & 3.4478 (p0.0486) & 4.0535 (p0.0331) & 3.4907 (p0.0669) \\
Truck count & Hackney – 1.014089941 & \( \alpha \) & 0.1311 (p0.7213) & 0.4454 (p0.6483) & 1.5225 (p0.2554) & 2.6202 (p0.0988) & 4.3103 (p0.0413) \\
 & (index=11) & \( \beta \) & 0.1518 (p0.6968) & 1.1693 (p0.5573) & 7.0267 (p0.0711) & 19.9132 (p0.0005) & 55.4185 (p0.0000) \\
 &  & \( \gamma \) & 0.1513 (p0.6973) & 1.1379 (p0.5661) & 6.0219 (p0.1106) & 13.6210 (p0.0086) & 25.3045 (p0.0001) \\
 &  & \( \delta \) & 0.1311 (p0.7213) & 0.4454 (p0.6483) & 1.5225 (p0.2554) & 2.6202 (p0.0988) & 4.3103 (p0.0413) \\
Bus (standing) & Richmond – 1.0662499 & \( \alpha \) & 13.2407 (p0.0017) & 6.7214 (p0.0076) & 4.7316 (p0.0192) & 4.2102 (p0.0297) & 1.9518 (p0.2037) \\
 & (index=420) & \( \beta \) & 15.3313 (p0.0001) & 17.6437 (p0.0001) & 21.8383 (p0.0001) & 31.9977 (p0.0000) & 25.0941 (p0.0001) \\
 &  & \( \gamma \) & 11.6334 (p0.0006) & 12.8071 (p0.0017) & 14.7616 (p0.0020) & 18.7595 (p0.0009) & 15.7143 (p0.0077) \\
 &  & \( \delta \) & 13.2407 (p0.0017) & 6.7214 (p0.0076) & 4.7316 (p0.0192) & 4.2102 (p0.0297) & 1.9518 (p0.2037) \\
Bus count & Hackney – 1.014070034 & \( \alpha \) & 2.1090 (p0.1628) & 1.5828 (p0.2359) & 1.6537 (p0.2256) & 1.6120 (p0.2459) & 6.4287 (p0.0150) \\
 & (index=9) & \( \beta \) & 2.4420 (p0.1181) & 4.1550 (p0.1252) & 7.6323 (p0.0543) & 12.2511 (p0.0156) & 82.6544 (p0.0000) \\
 &  & \( \gamma \) & 2.3157 (p0.1281) & 3.7912 (p0.1502) & 6.4650 (p0.0911) & 9.4547 (p0.0507) & 30.9838 (p0.0000) \\
 &  & \( \delta \) & 2.1090 (p0.1628) & 1.5828 (p0.2359) & 1.6537 (p0.2256) & 1.6120 (p0.2459) & 6.4287 (p0.0150) \\
Bus count & Hackney – 1.0141400098 & \( \alpha \) & 12.1979 (p0.0024) & 6.2016 (p0.0101) & 6.2687 (p0.0073) & 3.0672 (p0.0686) & 4.7238 (p0.0331) \\
 & (index=15) & \( \beta \) & 14.1239 (p0.0002) & 16.2792 (p0.0003) & 28.9323 (p0.0000) & 23.3110 (p0.0001) & 60.7350 (p0.0000) \\
 &  & \( \gamma \) & 10.9101 (p0.0010) & 12.0522 (p0.0024) & 17.8941 (p0.0005) & 15.2115 (p0.0043) & 26.5629 (p0.0001) \\
 &  & \( \delta \) & 12.1979 (p0.0024) & 6.2016 (p0.0101) & 6.2687 (p0.0073) & 3.0672 (p0.0686) & 4.7238 (p0.0331) \\
Congestion & Hackney – 1.01301002 & \( \alpha \) & 2.0070 (p0.1728) & 1.5517 (p0.2421) & 1.2760 (p0.3237) & 1.8226 (p0.2011) & 7.1196 (p0.0114) \\
 & (index=2) & \( \beta \) & 2.3239 (p0.1274) & 4.0732 (p0.1305) & 5.8893 (p0.1171) & 13.8517 (p0.0078) & 91.5375 (p0.0000) \\
 &  & \( \gamma \) & 2.2092 (p0.1372) & 3.7228 (p0.1555) & 5.1620 (p0.1603) & 10.4037 (p0.0341) & 32.5061 (p0.0000) \\
 &  & \( \delta \) & 2.0070 (p0.1728) & 1.5517 (p0.2421) & 1.2760 (p0.3237) & 1.8226 (p0.2011) & 7.1196 (p0.0114) \\
\hline
\end{tabular}
\label{tab:granger_causality_analysis}
\end{sidewaystable}


\begin{table*}[h]
\centering
\caption{The results of the t-test models for all factors grouped by whether the samples are outside or inside the ULEZ. Test 1: NO\textsubscript{2} outside and inside ULEZ, Test 2: Truck flow outside and in the ULEZ zone, Test 3: Bus flow outside and in the ULEZ zone, Test 4: Car flow outside and in the ULEZ zone, Test 5: Bicycle flow outside and in the ULEZ zone}
\scriptsize  
\begin{tabular}{lccccc}
\hline
\textbf{Independent t-test} & \textbf{Test 1} & \textbf{Test 2} & \textbf{Test 3} & \textbf{Test 4} & \textbf{Test 5} \\
\hline
\textbf{Difference (outside-inside the zone)} & -3.7439 & 2.22 & -3.1366 & 25.1771 & -1.2931 \\
\textbf{Degrees of freedom} & 88020 & 88020 & 88020 & 88020 & 88020 \\
\textbf{t} & -25.7615 & 31.6335 & -58.9364 & 59.0456 & -54.5541 \\
\textbf{Two side test p value} & 0 & 0 & 0 & 0 & 0 \\
\textbf{Difference < 0 p value} & 0 & 1 & 0 & 1 & 0 \\
\textbf{Difference > 0 p value} & 1 & 0 & 1 & 0 & 1 \\
\textbf{Cohen's d} & -0.1748 & 0.2146 & -0.3999 & 0.4006 & -0.3701 \\
\textbf{Hedge's g} & -0.1748 & 0.2146 & -0.3999 & 0.4006 & -0.3701 \\
\textbf{Glass's delta1} & -0.1821 & 0.1948 & -0.495 & 0.3489 & -0.5017 \\
\textbf{Point-Biserial r} & -0.0865 & 0.106 & -0.1948 & 0.1952 & -0.1808 \\
\hline
\end{tabular}
\label{tab:t_test_models}
\end{table*}


\begin{sidewaystable}
\centering
\caption{The results of trained models for inferring NO\textsubscript{2} levels. All results are shown for the validation set. All models are trained using Nvidia RTX A5000-16GB, CPU Xeon -12 cores, Ram 128G. Architectures: A: Graph Convolutional model, B: Attention-based Graph model, C: Transformer model, D: Conv1D-based multiple branch model. Task 1 represents NO\textsubscript{2} detection at a given location and Task 2 represents NO\textsubscript{2} detection at city-wide.}
\scriptsize  
\begin{tabular}{|l|c|c|c|c|c|c|c|c|c|c|c|}
\hline
\textbf{Task} & \textbf{Model ID} & \textbf{Signature (N)} & \textbf{Edge} & \textbf{Architecture} & \textbf{Params} & \textbf{Train time (h)} & \textbf{MSLE} & \textbf{MAE} & \textbf{MSE} & \textbf{Kl-divergence} & \textbf{R2} \\
\hline
\multirow{4}{*}{\rotatebox[origin=c]{90}{Task 1}} 
    & 1 & Signature (N=3) & KNN - 10 & A & 55,839 & 6.4 & 0.0375 & 0.6558 & 0.6558 & - & 0.41 \\
    & 2 & No Signature & KNN - 10 & A & 55,839 & 6.4 & 0.0281 & 0.5722 & 0.5722 & - & 0.21 \\
    & 3 & No Signature & KNN - 10 & B & 120,342,324 & 5.15 & 0.0454 & 0.6842 & 0.6930 & - & 0.40 \\
    & 4 & Signature (N=3) & KNN - 10 & B & 120,342,324 & 5.15 & 0.0517 & 0.7249 & 0.9386 & - & 0.34 \\
\hline
\multirow{16}{*}{\rotatebox[origin=c]{90}{TASK 2}} 
    & 5 & No Signature & KNN - 10 & C & 8,258,720 & 0.6 & 0.0526 & 0.7313 & 0.8097 & 0.0044 & 0.39 \\
    & 6 & No Signature & KNN - 50 & C & 8,258,720 & 0.6 & 0.0374 & 0.6019 & 0.5516 & 0.0032 & 0.35 \\
    & 7 & Signature (N=3) & KNN - 10 & C & 164,498,712 & 2.16 & 0.0498 & 0.7007 & 0.7511 & 0.0005 & 0.65 \\
    & 8 & Signature (N=3) & KNN - 50 & C & 164,498,712 & 2.16 & 0.0479 & 0.6836 & 0.7360 & 0.0004 & 0.73 \\
    & 9 & No Signature & KNN - 10 & D & 6,367,866 & 0.6 & 0.0380 & 0.6109 & 0.6444 & 0.0004 & 0.75 \\
    & 10 & No Signature & KNN - 20 & D & 6,367,866 & 0.6 & 0.0337 & 0.5751 & 0.5053 & 0.0003 & 0.79 \\
    & 11 & No Signature & KNN - 30 & D & 6,367,866 & 0.6 & 0.0370 & 0.6118 & 0.6012 & 0.0041 & 0.76 \\
    & 12 & No Signature & KNN - 40 & D) & 6,367,866 & 0.6 & 0.0309 & 0.5524 & 0.4752 & 0.0018 & 0.79 \\
    & 13 & No Signature & KNN - 50 & D & 6,367,866 & 0.6 & 0.0337 & 0.5668 & 0.5319 & 0.0181 & 0.78 \\
    & 14 & No Signature & KNN - 60 & D & 6,367,866 & 0.6 & 0.0331 & 0.5723 & 0.5125 & 0.0125 & 0.78 \\
    & 15 & Signature (N=3) & KNN - 10 & D & 7,909,588 & 0.8 & 0.0345 & 0.5782 & 0.5741 & 0.0003 & 0.75 \\
    & 16 & Signature (N=3) & KNN - 20 & D & 7,909,588 & 0.8 & 0.0308 & 0.5503 & 0.4644 & 0.00025 & 0.79 \\
    & 17 & Signature (N=3) & KNN - 30 & D & 7,909,588 & 0.8 & 0.0346 & 0.5842 & 0.550 & 0.0284 & 0.77 \\
    & 18 & Signature (N=3) & KNN - 40 & D & 7,909,588 & 0.8 & 0.0299 & 0.5426 & 0.4508 & 0.00001 & 0.81 \\
    & 19 & Signature (N=3) & KNN - 50 & D & 7,909,588 & 0.8 & 0.0285 & 0.5197 & 0.4251 & 0.00001 & 0.82 \\
    & 20 & Signature (N=3) & KNN - 60 & D & 7,909,588 & 0.8 & 0.0294 & 0.5377 & 0.4452 & 0.00001 & 0.82 \\
\hline
\end{tabular}
\label{tab:trained_models_no2_levels}
\end{sidewaystable}

\end{document}